\documentclass[journal]{IEEEtran}

\usepackage{subcaption}
\usepackage{caption}
\usepackage{graphbox}
\usepackage{comment}
\usepackage{multirow}
\usepackage{booktabs}
\usepackage{tabu}
\usepackage{tabularx}
\usepackage[table]{xcolor}
\usepackage{enumitem}
\usepackage{lscape}  
\usepackage{adjustbox}  
\usepackage{balance}
\usepackage{rotating}
\newcommand*\rot{\rotatebox{90}}
\usepackage{pifont}
\newcommand{\cmark}{\textcolor{black} {\ding{51}}}%
\newcommand{\xmark}{\textcolor{black!30} {\ding{55}}}%
\graphicspath{{figures/}{biographies/}}
\usepackage{picinpar}

\begin{document}
\title{Revisiting Crowd Counting: State-of-the-art, Trends, and Future Perspectives}


\author{Muhammad Asif~Khan, Hamid Menouar,
and~Ridha Hamila
\thanks{M. A. Khan and H. Menouar are with Qatar Mobility Innovations Center (QMIC), Doha, Qatar.}
\thanks{R. Hamila is with Qatar University.}
}

\markboth{Submitted to Journal of Image and Vision Computing}%
{Khan \MakeLowercase{\textit{et al.}}: Revisiting Crowd Counting: State-of-the-art, Trends, and Future Perspectives}

\maketitle  

\begin{abstract}
Crowd counting is an effective tool for situational awareness in public places. Automated crowd counting using images and videos is an interesting yet challenging problem that has gained significant attention in computer vision. Over the past few years, various deep learning methods have been developed to achieve state-of-the-art performance. The methods evolved over time vary in many aspects such as model architecture, input pipeline, learning paradigm, computational complexity, and accuracy gains etc. In this paper, we present a systematic and comprehensive review of the most significant contributions in the area of crowd counting. Although few surveys exist on the topic, our survey is most up-to date and different in several aspects. First, it provides a more meaningful categorization of the most significant contributions by model architectures, learning methods (i.e., loss functions), and evaluation methods (i.e., evaluation metrics). We chose prominent and distinct works and excluded similar works. We also sort the well-known crowd counting models by their performance over benchmark datasets. We believe that this survey can be a good resource for novice researchers to understand the progressive developments and contributions over time and the current state-of-the-art.
\end{abstract}


\section{Introduction} \label{sec:intro}
Crowd counting is an interesting research area with many real-world applications. The increasing population and growing urbanization trends often result in rapid crowds creation in urban areas such as metro stations, sports venues, musical concerts, exhibition centers, and parade grounds etc. The prediction of the crowds creation, and the density estimation in crowded regions is of paramount significance in crowd monitoring, management, and effective events planning. The social distancing measures to reduce the spread of virus observed during the recent COVID-19 pandemic further highlight its significance \cite{salma_2021, saad_2022}. Owing to the importance of the problem, a huge amount of research exists on automated crowd counting using image and video analysis methods. Although traditional image processing methods have shown limited performance, the last decade has witnessed major improvements using state-of-the-art methods in computer vision and deep learning.
\par
Crowd counting is generally implemented in two ways: (i) counting objects (input is an image and the output is a number i.e., total head count in the image.), and (ii) density map estimation (input is an image and the output is the density map of the crowd which is then integrated to get the total head count.). 

Traditional methods for crowd counting were all based on total count approach. These methods employ image processing techniques to detect hand-crafted features e.g., body appearance \cite{Lin_2010, Tuzel_2008, Leibe_2008}, or body parts \cite{Viola_2001, Lin_2001, Li_2008, Felzenszwalb_2010, Topkaya_2014, Wu_2006} and then use machine learning models such as linear regression, ridge regression, Gaussian process, support vector machines (SVMs), random forest, gradient boost, and neural networks to find the total head count in an image. However, all the accuracy of these methods significantly degrades on images with dense crowds due to challenges such as occlusions, low resolution, foreshortening and perspectives.

Due to the aforementioned limitations of detection-based approach, regression-based methods were adopted which regress the total count from an image or image patch. Instead of detecting body parts or shapes, regression-based methods use global features such as texture \cite{Chen_2012}, foreground \cite{Davies_1995}, gradients \cite{Dalal_2005, Chan_2009} etc. to estimate image-wise or patch-wise total count \cite{Paragios_2001, Tian_2010}. Regression-based method solve the problems raised from occlusions, low resolution, and perspective. However, regression-based method show poor performance on high density crowd images.
\par
Recent research on crowd counting shows the efficacy of convolution neural networks (CNNs) \cite{MCNN_CVPR2016, CSRNet_CVPR2018, SANet_ECCV2018, PGCNet_ICCV2019, ASNet_CVPR2020, SGANet_IEEEITS2022} due to their strong capability of automatic feature extraction. Like many other computer vision problems, such as image classification \cite{VGG16_ICLR2015, Inception_CVPR2015}, object detection \cite{YOLO_CVPR2016}, and image segmentation \cite{SegNet_2015, UNet_2015}, CNN is used as a widely applied method for crowd counting and it outperforms all the traditional methods. While traditional methods for crowd counting predict the total head count in an image, CNN-based methods often employ crowd density estimation. Density estimation is when the CNN model is used to predict the density map of the crowd scene rather than just the head count. The density map contains additional location information along with the total head count of the crowd scene.

First introduced in \cite{CrowdCNN_CVPR2015}, density estimation using CNNs has been adopted in all subsequent research contributions. However, there have been major architectural improvements to achieve optimum performance. As the performance of any deep learning method typically is evaluated using benchmark datasets, many crowd datasets were also published over time. These datasets published over time introduced more challenges in terms of high density, scale variation, scene variation, and illumination changes, crowd distribution, extreme occlusions, perspective distortions etc. To overcome these challenges and to achieve high accuracy over these datasets, complex CNN architectures, novel learning methods, and sophisticated evaluation criteria were developed over time.
This survey covers all these developments so that a novice researcher in the area of crowd counting can understand these advancements in an incremental manner. Although there exists few relevant surveys \cite{Saleh_2015, Jeevitha_2018, SnehaPK_2018, Sindagi_2018, Cenggoro_2019, Naveed_2020, Gao_2020, Abdou_2020, Luo_2020, Bai_2020, Jingying_2020, Gouiaa_2021, Fan_2022}, we believe our work is different in several aspects:
\begin{itemize}
\item There has been several significant contributions over the last 1-3 years which are not covered in the old surveys \cite{Saleh_2015, Jeevitha_2018, SnehaPK_2018, Sindagi_2018, Cenggoro_2019, Naveed_2020, Gao_2020, Abdou_2020, Luo_2020, Bai_2020, Jingying_2020}. These new contributions span multiple directions including novel and improved loss functions, more accurate evaluation metrics, and novel network architectures (e.g., transformer-based models). Thus, this survey is a more up to date reference on the topic.
\item We provide a highly intuitive review of the state-of-the-art by more meaningful segregation of the literature so that each contribution is discussed separately. For instance, the segregation of works by network architecture, loss functions, training methods etc. and then sorting by performance is of more interest to the novice researchers as well as general readers. 
\item Unlike other surveys, we intend to focus our discussion on the most significant contributions based on their impact and the relative performance gains over previous works. By excluding redundant and least significant contributions (in terms of novelty or performance), our survey is concise, more focused and more readable to follow the actual advancement of the state-of-the-art rather than repeatedly discussing similar ideas of least impact.
\end{itemize}

The rest of the paper is organized as follows:
We first discuss (in Section \ref{sec:challenges}) the various challenges of crowd counting with a visual illustration. Then a brief overview of crowd datasets used in benchmarking of previous research works is provided in Section \ref{sec:datasets}. Section \ref{sec:annotation} explains the data annotation techniques applied in crowd counting and density estimation. Section \ref{sec:models} presents a comprehensive review, critical analysis, and performance comparison of crowd counting models. Section \ref{sec:train_eval} presents an indepth analysis of training and evaluation methods for crowd counting covering all the loss functions and evaluation metrics used in the literature. Lastly, we concluded the paper in Section \ref{sec:conclusion} with some interesting insights for prospective researchers working in this area.

\section{Crowd Counting Challenges} \label{sec:challenges}
A robust crowd counting model aims to accurately predict crowd count and estimate local as well as global density in crowd scenes under diverse conditions. The varying conditions heavily impact the performance of the counting model. Hence, it is therefore important to first understand these challenges and their possible impact on the model performance. A good understanding of these challenges helps in developing more robust models. Some of the more common challenges in crowd counting are listed here with a visual illustration in Fig. \ref{fig:crowd_challenges}.

\begin{figure*}[!h] \label{fig:crowd_challenges}
\centering

\subcaptionbox{\scriptsize Single Scene dataset.}{
\includegraphics[width=3cm, height=2cm]{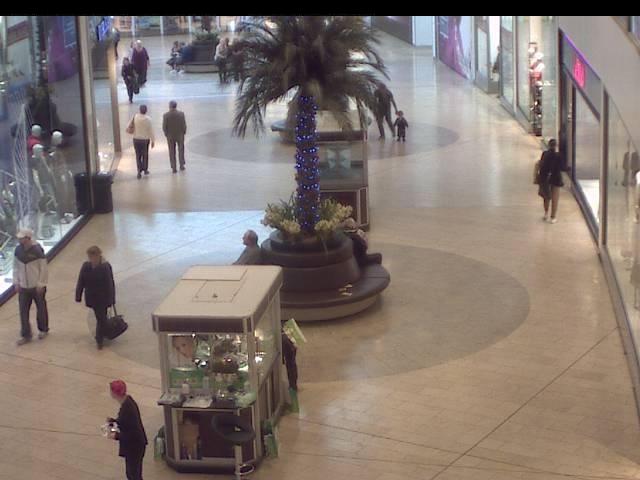}
\includegraphics[width=3cm, height=2cm]{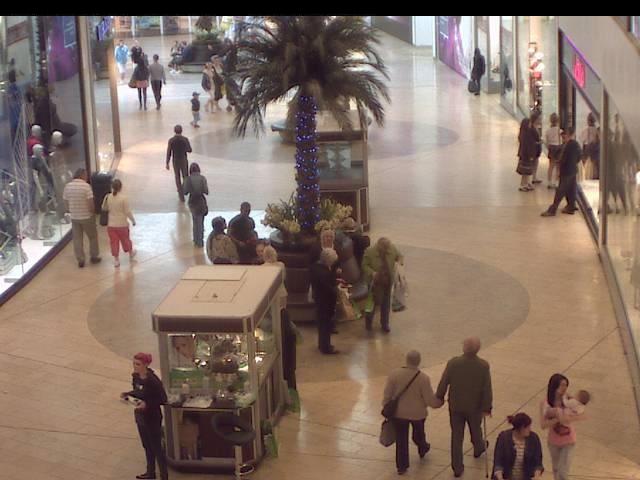}}
\hspace{2em}
\subcaptionbox{\scriptsize Multiple scenes dataset.}{
\includegraphics[width=3cm, height=2cm]{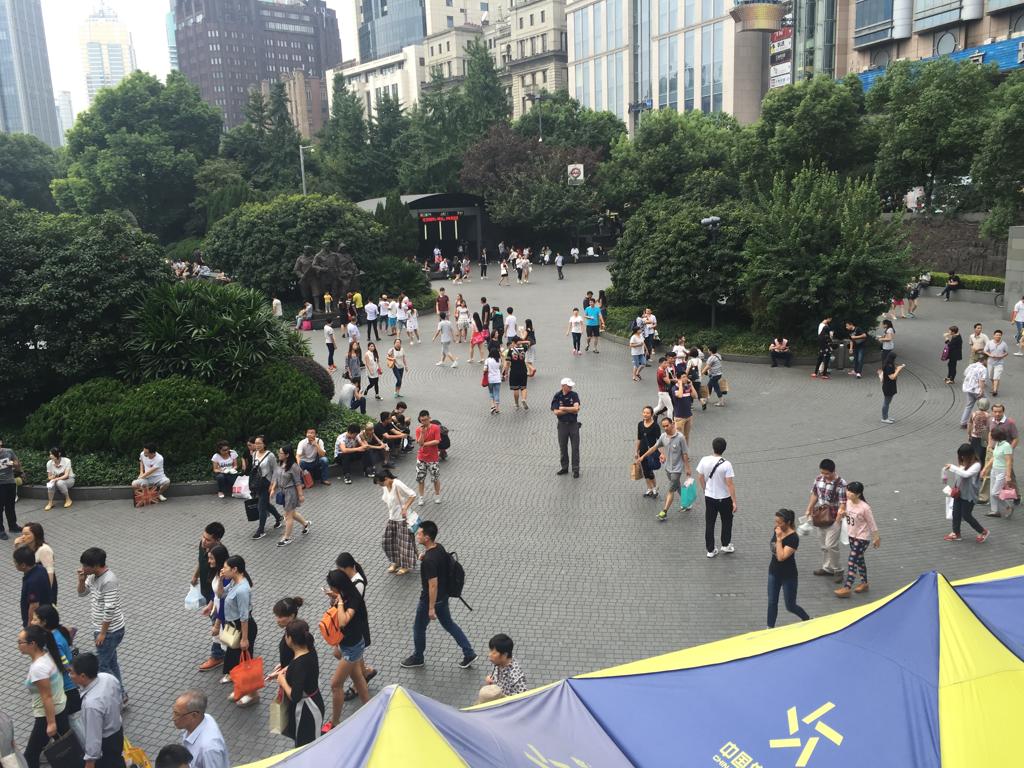}
\includegraphics[width=3cm, height=2cm]{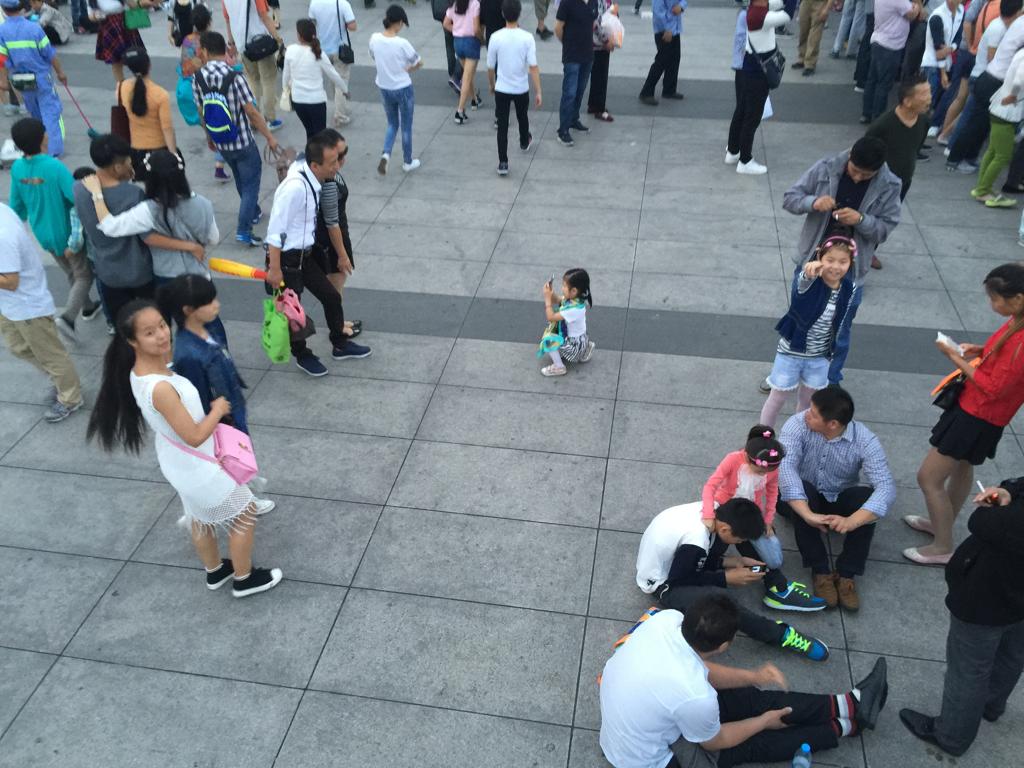}}
\\ \vspace{1em}

\subcaptionbox{\scriptsize Scale variation in same image.}{
\includegraphics[width=3cm, height=2cm]{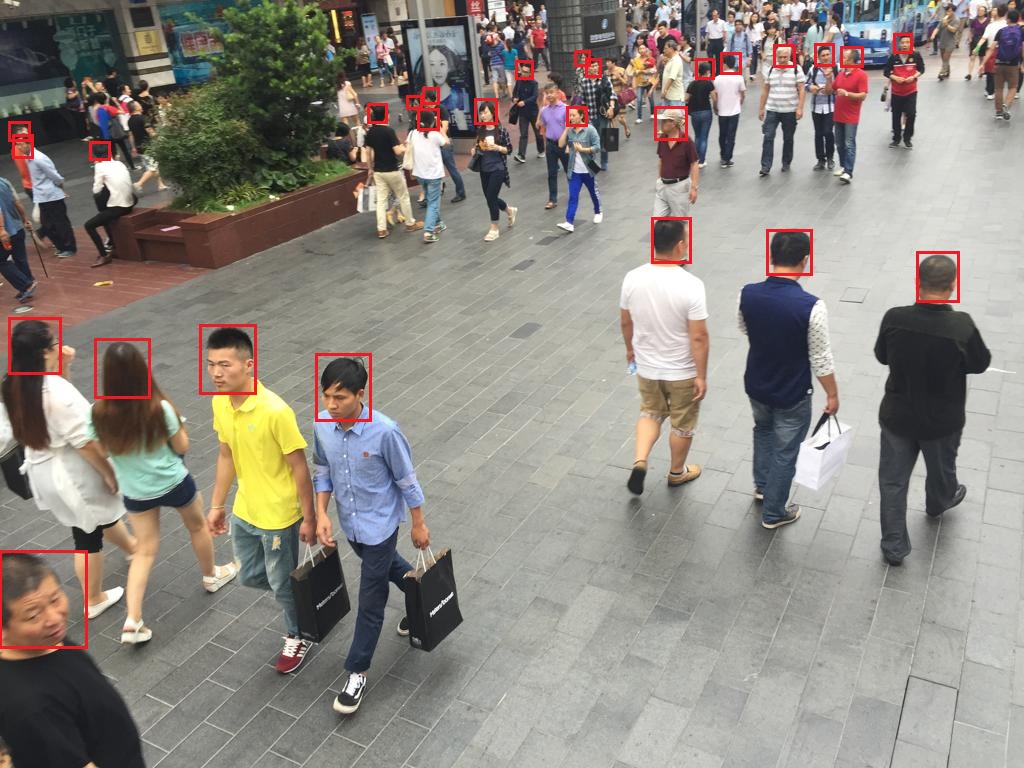}
\includegraphics[width=3cm, height=2cm]{figures/images/scale1.jpg}}
\hspace{2em}
\subcaptionbox{\scriptsize Scale variation in different images of the same scene.}{
\includegraphics[width=3cm, height=2cm]{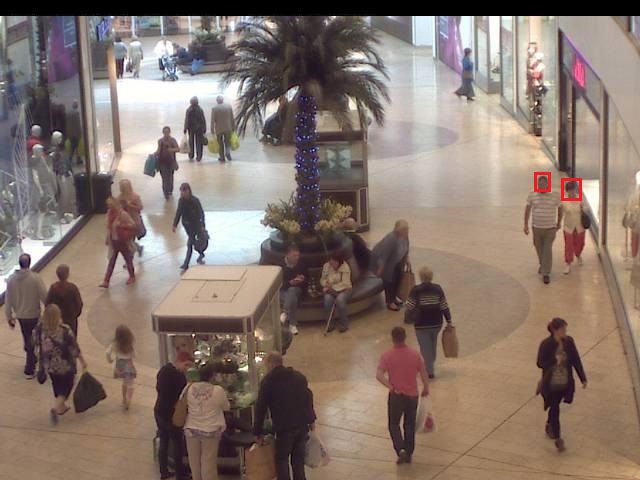}
\includegraphics[width=3cm, height=2cm]{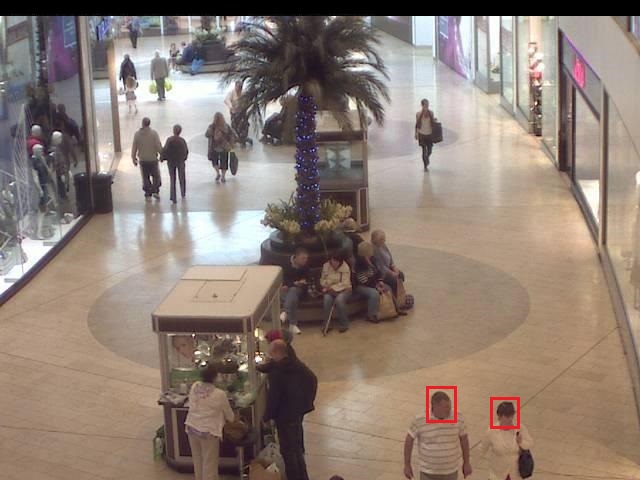}}
\\ \vspace{1em}

\subcaptionbox{\scriptsize Variations in crowd density (low vs high density).}{
\includegraphics[width=3cm, height=2cm]{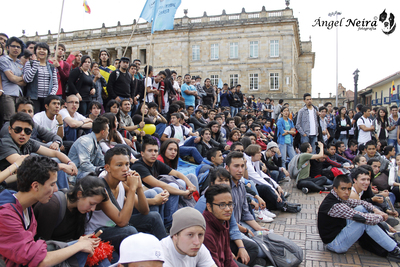}
\includegraphics[width=3cm, height=2cm]{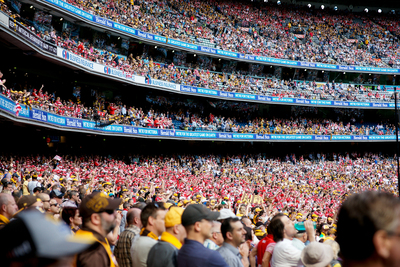}}
\hspace{2em}
\subcaptionbox{\scriptsize Uniform vs non-uniform crowd distribution.}{
\includegraphics[width=3cm, height=2cm]{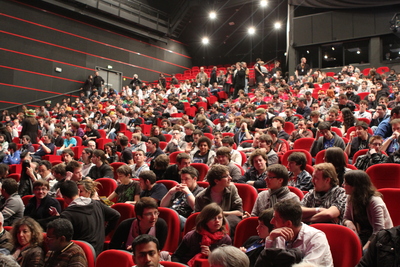}
\includegraphics[width=3cm, height=2cm]{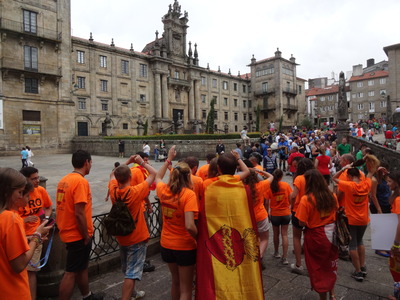}}
\vspace{1em}

\subcaptionbox{\scriptsize Occlusions in crowd images.}{
\includegraphics[width=3cm, height=2cm]{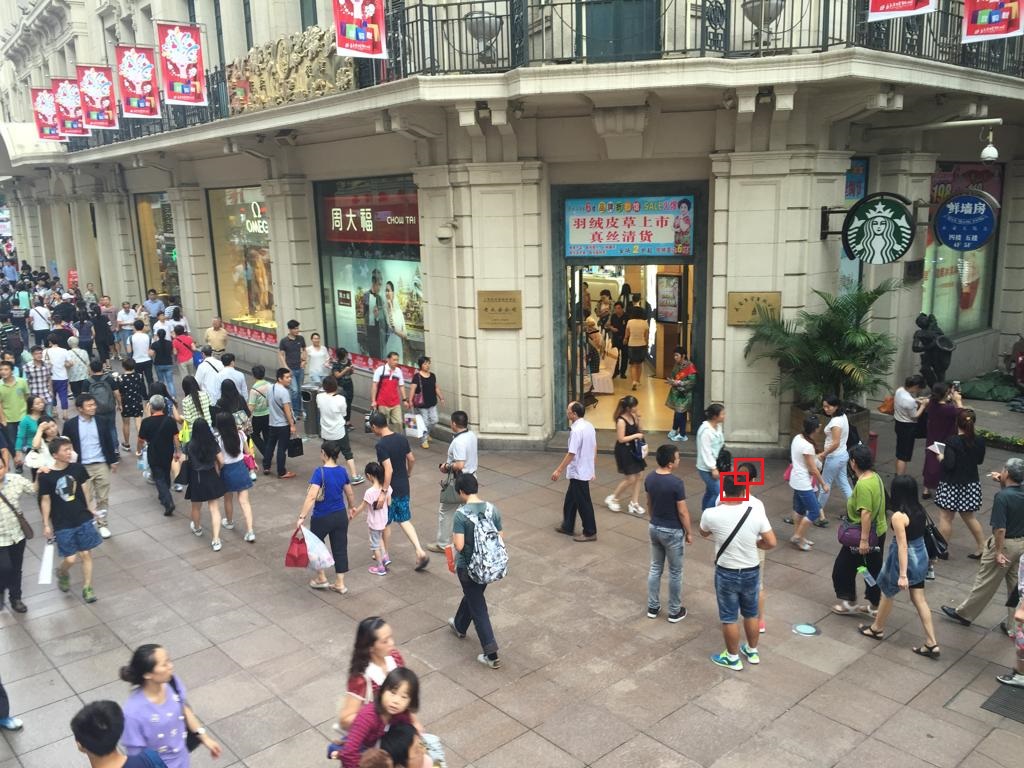}
\includegraphics[width=3cm, height=2cm]{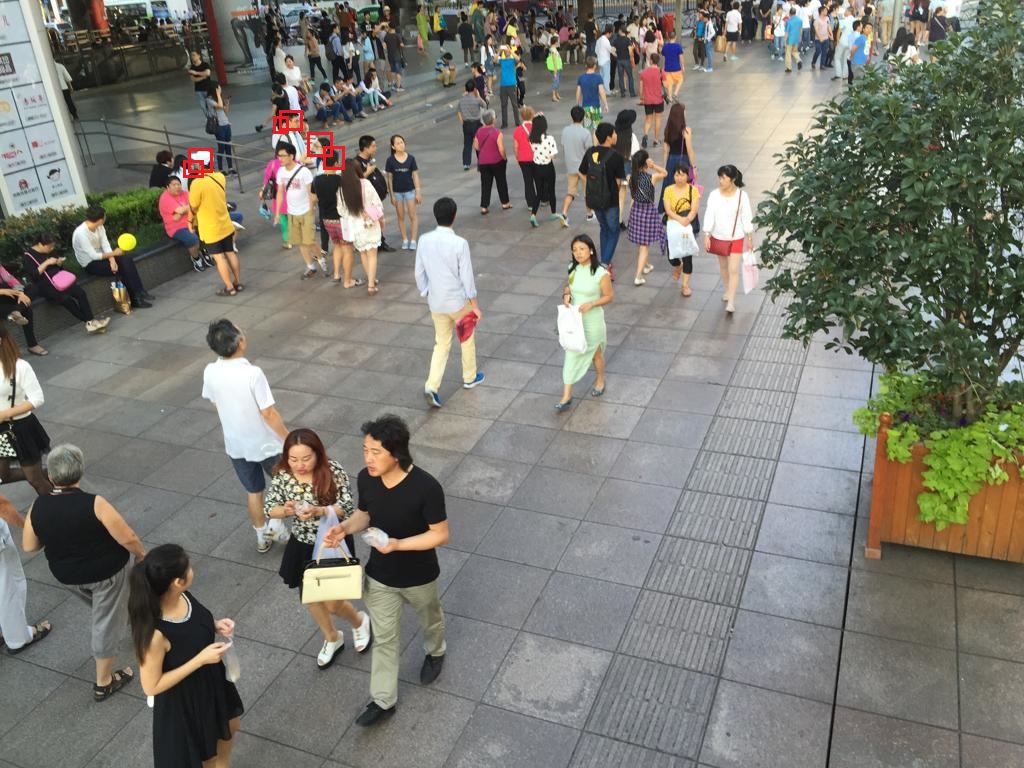}}
\hspace{2em}
\subcaptionbox{\scriptsize Complex backgrounds.}{
\includegraphics[width=3cm, height=2cm]{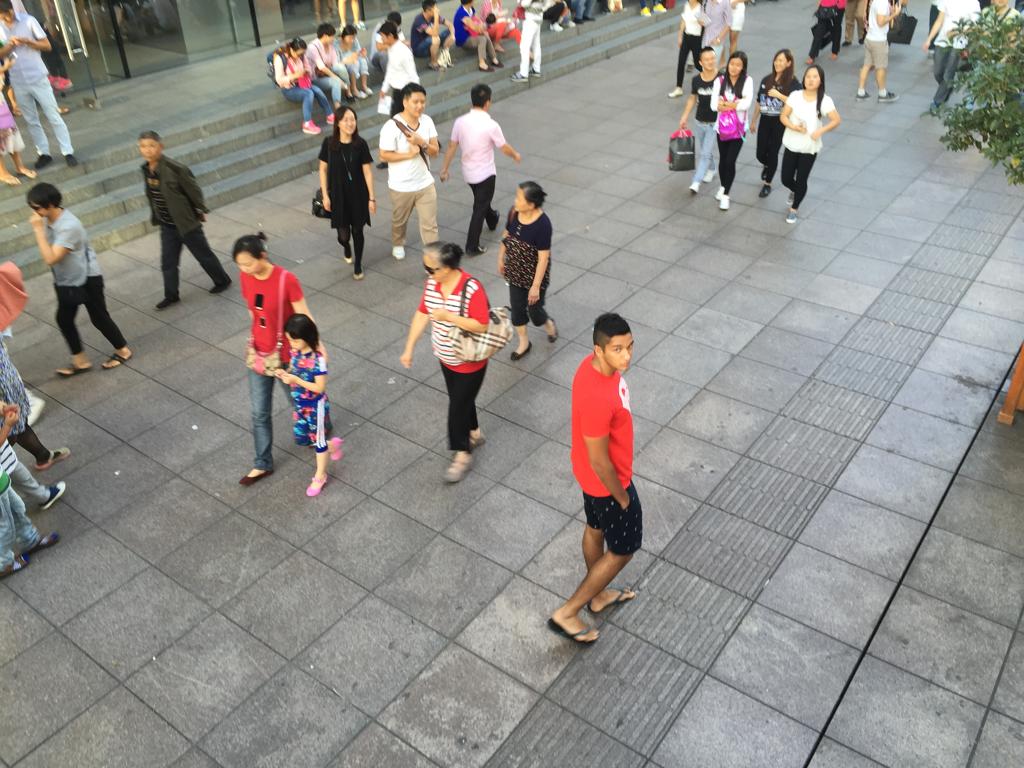}
\includegraphics[width=3cm, height=2cm]{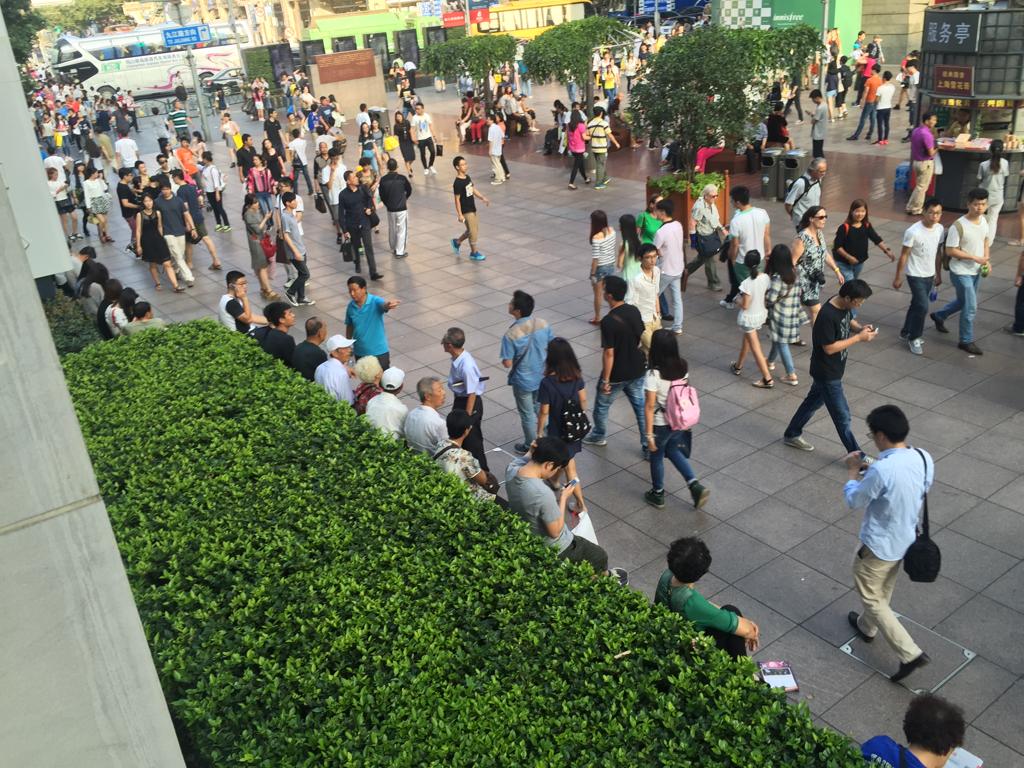}}

\caption{Challenges in crowd counting using computer vision.} \label{fig:crowd_challenges}
\end{figure*}

\begin{enumerate}
\item  \textit{Scene variation:}
Crowd monitoring across different scenes is more challenging. For instance, counting people in images taken by a single CCTV camera (e.g., in Mall dataset \cite{Mall_dataset2012}) is much easier than counting people in images taken by several surveillance cameras (e.g., in WorldExpo'10 dataset \cite{CrowdCNN_CVPR2015}). Scene variations is common in drone surveillance and becomes more challenging when combined with other variations listed afterwards.

\item \textit{Scale Variation:}
Scale variation refers to the situation when objects related to the same class (i.e., humans) appear in different sizes in a single image as well as in different images. Scale variation is caused by distance (between camera and objects) and perspective effect in the same image. Moreover, scale variations are also observed in images of different resolutions. Scale variation is one of the most common challenges that hugely impact the model performance and is frequently addressed in crowd counting research.

\item \textit{Crowd Density:}
The number of people (or other objects of interests) vary from one image to another. Usually, images with low density are easy than highly dense images. Another challenge is when the same image contains different crowd densities in different regions.
    
\item \textit{Non-uniform People Distribution:}
Crowd images may differ in terms of distribution of objects (e.g., people). For instance, people sitting in a sports stadium are uniformly distributed with constant distance among objects. In contrast, objects may be distributed randomly in street crowds. Uniformly distributed crowds can be estimated more accurately than non-uniform crowds in the absence of other attributes affecting the accuracy.

\item  \textit{Occlusions:}
Occlusion refers to the overlap in objects. When similar objects (e.g., people) overlap, it is called intra-class occlusion, whereas when different objects (e.g., people, cars, walls) overlap, it is called inter-class occlusion. Dealing with occlusion is often challenging. It is not only difficult for annotators to annotate the objects in the presence of occlusion, but also challenging for the object detectors to accurately predict objects. Occlusion interweave semantic features in images making it difficult to discriminate object boundaries.

\item \textit{Complex background:}
The background region which do not contain objects of interest vary in the same images as well as across different images. When the pixel values in the background have similar values as that of the object, it can make learning hard.

\item \textit{Illumination variations:}
The illumination in an image vary at different times of a day and may vary in different regions in the same images due to lighting conditions. Thus the same object (e.g., people) in the same image will have different pixel values, which makes learning challenging.

\item \textit{Other variations:}
The aforementioned attributes in crowd images is not an exhaustive list. There are several other attributes that make counting a more challenging problem. Few examples are weather changes, image distortion and noise, object rotation etc.
\end{enumerate}

\section{Crowd Counting Datasets} \label{sec:datasets}
A number of benchmarking datasets have been introduced over time that are being used for evaluating crowd counting models. These datasets vary in size (number of samples), annotations, and image attributes. Table \ref{tab:crowd_datasets} provides a brief summary of these datasets.

\begin{table*}[!h]
\centering \caption{Crowd counting datasets.}
\small
\begin{tabular}{r|c|c|c|c} \toprule
Dataset    &Attribute     &Scene               &Images  &Resolution 
\\ \midrule \midrule
UCSD \cite{UCSD_dataset2008}             &CCTV  &Single   &2000     &$238\times158$ \\

Mall \cite{Mall_dataset2012}              &CCTV  &Single   &2000     &$640\times480$ \\

UCF\_CC\_50 \cite{UCF_CC_50_dataset2013}       &Free  &Single   &50 &Vary \\

WorldExpo'10 \cite{CrowdCNN_CVPR2015}     &Video frames  &Cross-scene   &3920     &$576\times720$ \\

ShanghaiTech-A \cite{MCNN_CVPR2016}    &Internet  &Single   &482     &Vary \\

ShanghaiTech-B \cite{MCNN_CVPR2016}    &CCTV  &Single   &716     &$768\times1024$ \\

UCF-QNRF \cite{CompositionLoss_2018}        &Free-view  &Single   &1535     &Vary \\

JHU-Crowd \cite{jhucrowd_dataset2020}        &Internet  &Cross-scene   &4372 &$1450\times900$ \\

DroneRGBT \cite{DroneRGBT_dataset}        &Drone  &Cross-scene   &3600     &$512\times640$ \\

NWPU Crowd \cite{NWPUCrowd_dataset2021}       &Internet  &Cross-scene   &5109    &Vary \\

TRANCOS  \cite{TRANCOS_dataset2015}         &CCTV  &Single   &1244     &$640\times480$ \\

CARPK  \cite{CARPK_dataset}           &Drone  &Cross-scene   &1448     &$1280\times720$ \\

VisDrone-People \cite{VisDrone_2021}  &Drone  &Cross-scene   &3347     &Vary \\

VisDrone-Vehicles \cite{VisDrone_2021} &Drone  &Cross-scene   &5303     &Vary \\

RGBT-CC \cite{RGBTCC_dataset2021} &Drone &Cross-scene &- &- \\

\bottomrule
\end{tabular}
\label{tab:crowd_datasets}
\end{table*}

\section{Data Labelling} \label{sec:annotation}
Crowd monitoring applications mostly use supervised learning or semi-supervised learning and thus require labelled data (images or videos). There are mainly three types of data annotations used for crowd images.

\subsection{Types of Annotation}

\begin{enumerate}

\item \textit{Total count annotation:} The total number of objects in the image is counted as a single number. This kind of annotation is used in detection-based and regression-based counting approaches.

\item \textit{Dot annotation:} The objects inside images are annotated as a single dot over each object, typically on the person's head. The dot annotation creates a dot-map (also called as localization map) i.e., a matrix consisting of all zeros except those positions containing heads in the image. All pixels corresponding to head positions are activated i.e., set to 1 per head. Dot annotation is used in density estimation methods for crowd counting. However, as the sum of dot map is equal to the total count in an image, it can also be used in count regression models.

\item \textit{Bounding box:} A rectangular bounding box is drawn around each object (e.g., person) in the image. Total count annotation is cheaper; however, it does not provide a good accuracy due to issues such as object occlusions. Dot annotation can outperform the total count, but it does not capture information such as object scaling, and it provide limited accuracy when the objects scale largely vary in images.

\subsection{Density-map generation}
Dot annotations is the most common and a relatively cheaper (than bounding box) annotation technique. However, the dot map or localization map which contains all zeros except head positions, are extremely sparse and it is generally very hard to train neural network over them \cite{CompositionLoss_2018}. Thus, in crowd counting research, such localization maps are converted to density maps.
If $x_i$ is a pixel containing the the head position, it can be represented by a delta function $\delta(x - x_i)$. The density map is generated by convolving the delta function with a Gaussian kernel $G_\sigma$.

\begin{equation}
    Y = \sum_{i=1}^{N}{ \delta(x-x_i) * G_\sigma}
\end{equation}
where, N is the total number of annotated points (i.e., total count of heads) in the image. The integral of density map $Y$ is equal to the total head count in the image.

\begin{figure} \centering
\includegraphics[width=0.9\columnwidth]{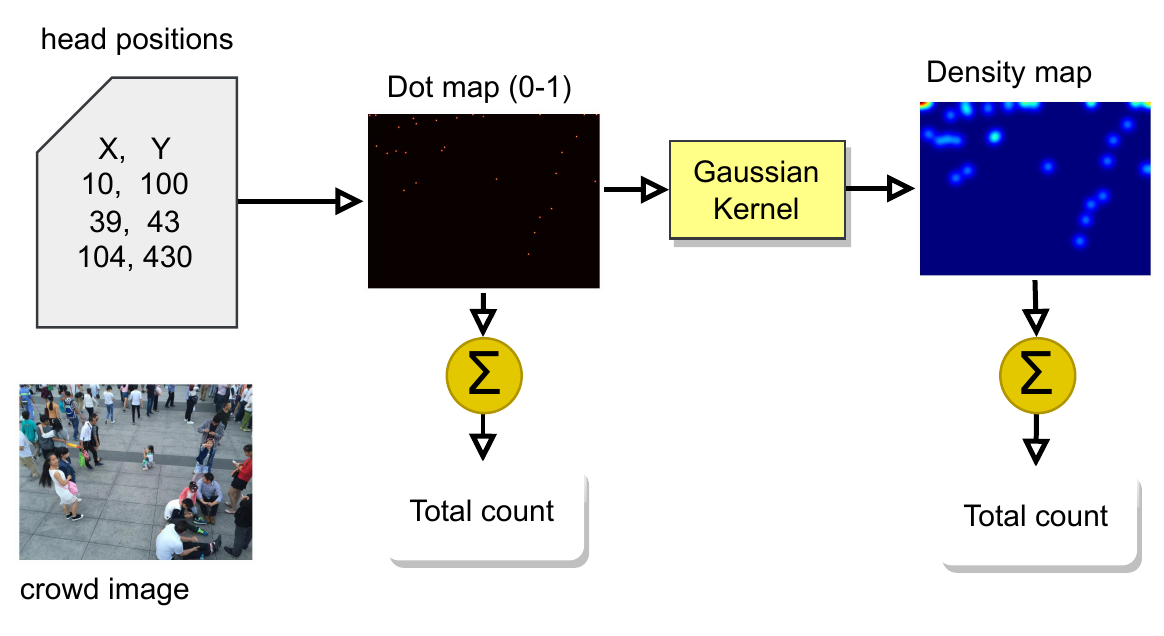}
\caption{Ground Truth generation for crowd counting models.} \label{fig:annotation}
\end{figure}

Visually, this operation creates a blurring of each head annotation using the scale parameter $\sigma$. There are various kernel settings to generate a variety of density maps.

The most basic approach is to keep $\sigma$ fixed value, which means that the density map will apply same kernel to all head positions irrespective of their scale in the image. Typical values of $\sigma$ used in earlier works are 15 \cite{SANet_ECCV2018}, X \cite{CrowdCNN_CVPR2015}.


\begin{table*}[]
\centering
\caption{Gaussian Kernels used in various studies.}
\small
\label{tab:fixed_kernels}
\renewcommand{\arraystretch}{0.5}
\begin{tabular}{r|p{5cm}|c|c} \toprule  
Model                   &Dataset   & Kernel &Scale ($\sigma$)  \\
\midrule \midrule
CrowdCNN \cite{CrowdCNN_CVPR2015} & -    &Adaptive &- \\[0.3em]
MCNN \cite{MCNN_CVPR2016} & -     &Adaptive &-\\[0.3em]
MSCNN \cite{MSCNN_ICIP2017} & -     &Adaptive &-\\[0.8em] 

\multirow{4}{*}{CSRNet \cite{CSRNet_CVPR2018}} &ShanghaiTech Part B & Fixed &15 \\
&TRANCOS  &Fixed &10\\
&UCSD, WorldExpo'10  &Fixed &3\\[0.5em]

SANet \cite{SANet_ECCV2018} & -   &Adaptive & -\\[0.3em]
TEDnet \cite{TEDnet_CVPR2019} & -   &Fixed & -\\[0.8em]

MobileCount \cite{MobileCount_PRCV2019} &ShanghaiTech Part B, UCF-QNRF, UCF-CC-50, WorldExpo'10  &Fixed &4\\[0.8em]

Deepcount \cite{DeepCount_ECAI2020} & -        &Fixed  &5 \\[0.3em]
BL \cite{BL_ICCV2019} &ShanghaiTechB    &Fixed &15 \\[0.3em]
C-CNN \cite{CCNN_ICASSP2020} &ShanghaiTech Part A \& B    &Fixed &15 \\[0.3em]
ASNet \cite{ASNet_CVPR2020} &-    &Fixed &- \\[0.3em]

\bottomrule
\end{tabular}
\end{table*}

Fixed Gaussian kernels are not helpful to learn the scale variations in images. Hence, adaptive kernels are proposed. In adaptive kernels, the value of $\sigma$ can be calculated as the average distance to k-nearest neighboring head annotations. Visually, it generates lower degree of Gaussian blur for dense crowds and higher degree for region of sparse density in crowd scene. Typical settings includes $k=10$ \cite{SANet_ECCV2018}, $k=10$ \cite{MSCNN_ICIP2017}.


\begin{figure}
\centering
\includegraphics[width=0.4\textwidth]{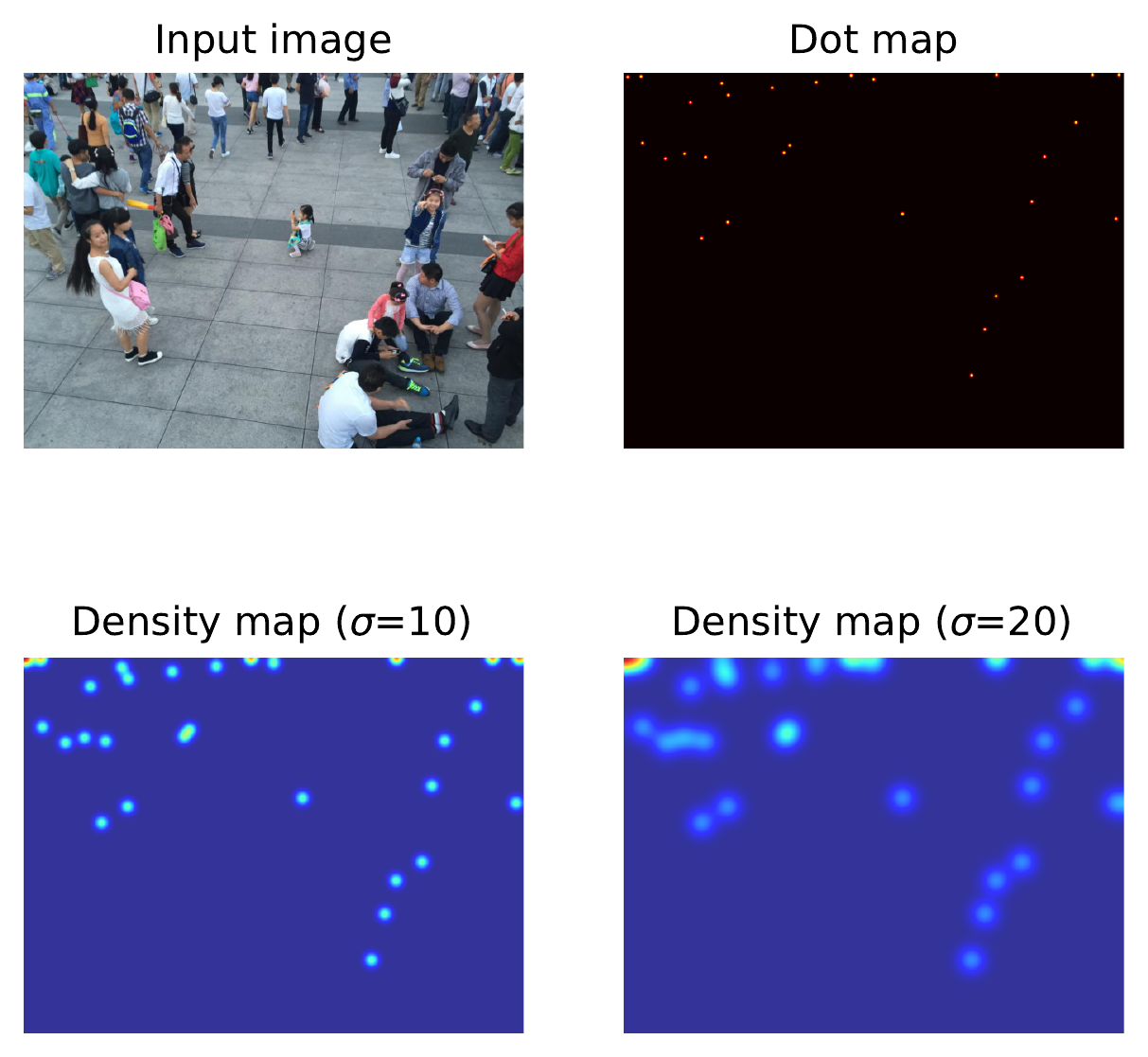}
\caption{Density maps with different scale ($\sigma$) values.}
\label{fig:gt_generation}
\end{figure}

\end{enumerate}

\subsection*{Summary and Lessons Learned}
\begin{itemize}
\item Fixed Gaussian kernels are adopted in earlier models because these are computationally cheaper than adaptive kernels. The value of $\sigma$ is often pre-determined empirically and can vary across different datasets. Interestingly, different values of $\sigma$ have been used for the same dataset in different works which potentially indicate the value may be fine-tuned for the CNN model to get better results.

\item Adaptive Gaussian kernels are increasingly used to cope with the scale variations in images. Earlier models used pre-determined values calculated as the weighted inter-head distances using K-nearest neighbors. The value of $k$ depends upon the crowd density i.e., higher value of $k$ for dense images and vice versa.

\item Some recent works propose to learn adaptive kernels alongside model training (explained in Section \ref{sec:train_eval}).
\end{itemize}


\section{Crowd Counting Models} \label{sec:models}
A wide numbers of DNN models are proposed for crowd counting over the period. These generally vary in size as well as design structure. We studied these models and categorized them into three distinct design categories i.e., (i) single-column, (ii) multi-column, and (iii) hybrid. While one may argue that such design patterns are common in other CNN models, it is important to understand the structure of these models in the context of crowd counting.

\subsection{Single-column Models}
One of the early models for crowd counting using CNN is reported in CrowdCNN \cite{CrowdCNN_CVPR2015}. The CrowdCNN model consists of three  convolution (Conv) layers (with kernels of $7 \times 7$, $7 \times 7$, and $5 \times 5$ respectively) followed by three fully connected (FC) layers. The input to the model is $72 \time 72$ patches cropped from image, whereas the output is a density map of $18 \times 18$ size. The model is evaluated over UCSD and UCF\_CC\_50 datasets using mean absolute error (MAE) as evaluation metric.
Early CNN models for crowd counting require perspective information during training as well as for inference, which may not be available in practical scenarios. In the absence of perspective maps, there may be significant performance degradation. Hence, very deep single-column models (e.g., FusionCount \cite{FusionCount_2022}), multi-column models, or single-column models with transfer learning are proposed over time. In \cite{SaCNN_WACV2018}, authors propose a single-column CNN (SaCNN) architecture with seven (7) Conv layers all using a single filter size ($3\times3$). The model uses explicitly $1\times1$ convolutions before $1\times1$ for dimension reduction and does not employ any pre-trained front-end CNN. The SaCNN performs multi-task learning i.e., simultaneously predict density map and head count. In \cite{BL_ICCV2019}, authors used a two-layer with kernel size of $3x3$ using VGG-19 as backbone. In compact-CNN (C-CNN) \cite{CCNN_ICASSP2020}, authors used three parallel filters of different size producing different outputs, which are merged and fed to five-layer single column CNN network.
Other single-column architectures include LibraNet \cite{LibraNet_ECCV2020}.

\subsection{Multi-column Models}
Single-column models usually are not good at detecting scale variations in object sizes. Multi-column approaches are favoured to learn multi-scale and perspective-free feature detection capability. The first multi-column CNN (MCNN) is proposed in \cite{MCNN_CVPR2016}. MCNN is a three-layers CNN architecture with difference receptive fields in each column. Each column in MCNN produces a density map of the same shape as ground truth map. The individual column's outputs are then concatenated to produce the final density map. MCNN takes an image of arbitrary size. Several works followed similar design pattern. For instance, the CrowdNet \cite{CrowdNet_CVPR2016} model proposes  a two-column CNN architecture. The first column is five-layer \textit{deep network} whereas the second column is a three-layer \textit{shallow network}. Similar to MCNN, the outputs of both columns are concatenated to predict the final density map. One of the drawback of multi-column models is the computation overhead. To dead with this, the Switch CNN (SCNN) \cite{SCNN_CVPR2017} model propose a 3-column network with different sized kernels in each column. However, an additional CNN network called classifier is used to automatically select only one column at inference time based on the crowd density.

A relatively simpler two-column network called cascaded multi-task learning (CMTL) is proposed in \cite{CMTL_AVSS2017} which jointly learns the density map as well as count density in images. In the two-column architecture, the first column predicts a high-level prior (i.e., the total head count) whereas the second column estimates the density map. Authors in \cite{DecideNet_CVPR2018} propose a detection and density estimation network (DecideNet) to jointly estimate count and density map. It consists of three columns however, unlike MCNN \cite{MCNN_CVPR2016}, each column in DecideNet performs a different task. The first column \textit{RegNet} is a 5-layer CNN which predicts the density map in the absence of target density map. The second column \textit{DetNet} uses Faster R-CNN network \cite{FasterCNN_ICCV2015} which first predicts the bounding box and then generates a detection-based density map. The third column \textit{QualityNet} uses the two density maps generated by RegNet and DetNet, and predicts the final density map.
In \cite{PACNN_CVPR2019}, authors propose a four-column CNN network with only the first four layers of VGG-16 as backbone. The first three columns generates density maps. Due to different size of filters in the three columns, when their outputs (density maps) are combined, it can adapt to the scale variation in person size. The fourth layer generates two perspective map which are used by the first two layers as intermediate inputs.
The DeepCount \cite{DeepCount_ECAI2020} architecture uses a large number of Conv layers and additionally introduces the intermediate fusion of layers at different stages.
A different architecture is proposed in ASNet \cite{ASNet_CVPR2020} which starts as a single-column (first seven layers), and then branch out into two columns. The first column produces density map, whereas the second column generates the scaling factor to be used in density estimation.
\par
Recently, multi-column models are increasingly used for multi-task learning. In PCCNet \cite{PCCNet_IEEE_TCSVT2020}, authors use a three column network to jointly learn the density map, density class, and segmentation map in crowd images. The three columns share encoded features at various stages.
Other multi-column architectures include MATT \cite{MATT_Elsevier_PR2020}, \cite{Yang_ICCV2020}, TAFNet \cite{TAFNET_2022}.

\subsection{Encoder-Decoder Models}
Dense CNN networks (e.g., VGG \cite{VGG16_ICLR2015}, Inception \cite{Inception_CVPR2015}, ResNets \cite{ResNet_CVPR2016}) designed for classification encode low-level spatial features into high-level semantic information. The down sampling operations used in these models causes loss of pixel-level localization information while encoding features. To generate the density map, the decoder layers aggregates the encoded feature maps to produce the final density map.
TEDnet uses the encoder-decoder architecture with hierarchical feature aggregation at different encoding stages. Both the encoder and decoder blocks uses multi-path encoding and decoding modules. The encoder block in TEDnet \cite{TEDnet_CVPR2019} uses the same multi-scale module as in Fig \ref{fig:SAM_SANet} which are originally proposed in \cite{SANet_ECCV2018}. Nine (9) multi-scale modules are cascaded with $2\times2$ pooling after the first two modules only. In the decoder block, multi-path decoding modules are used to aggregate the features and restore the spatial resolution in the predicted density maps. The network loss in TEDnet is computed at three intermediate levels in the decoder block, and one at the final output of the model.
A lightweight encoder-decoder architecture is MobileCount \cite{MobileCount_PRCV2019} designed for computational efficiency. The encoder block consists of first four layers of MobileNet\_V2 \cite{MobileNetV2_CVPR2018} and decoder block based on RefineNet \cite{RefineNet_2018}.
In \cite{ANF_CVPR2019}, authors propose a compact encoder-decoder model called attentional neural fields (ANF). ANF uses conditional random fields (CRFs) to aggregate multi-scale features inside encoder-decoder model. Both the encoder and decoder blocks have six (6) Conv layers with both inter-layer and intra-layer attention mechanism to capture dependencies among features of same scale and across different scales. Other examples include SASNet \cite{SASNet_AAAI2021}, MFCC \cite{MFCC_2022}.

\subsection{Single-column Models with Multi-column modules}
\begin{figure}[!h] \centering
\includegraphics[width=0.9\columnwidth]{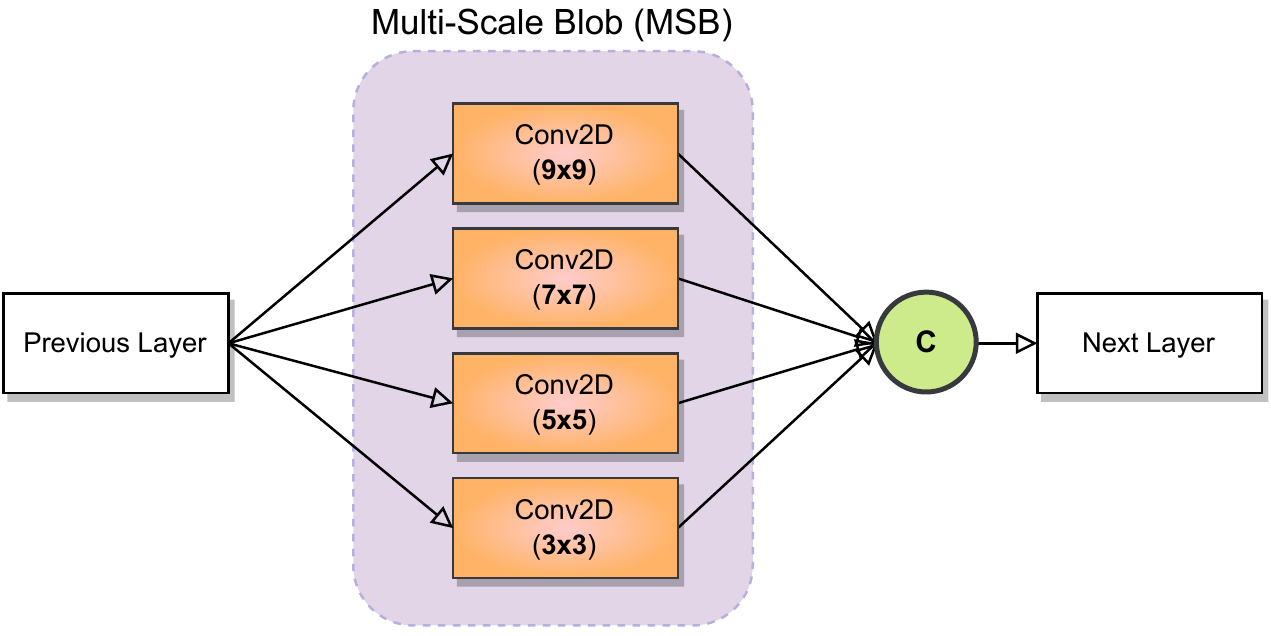}
\caption{Inception \cite{Inception_CVPR2015} like multi-scale blob (MSB) used in MSCNN \cite{MSCNN_ICIP2017}.} \label{fig:MSB_MSCNN}
\end{figure}

This type of network architecture has been used in many works mainly to cope with the shortcomings of both single-column and multi-column modules. Single column models are less complex but perform poorly in detecting multi-scale features wheres multi-column can detect multi-scale features but their capability is limited by the number of columns. The fact that adding more columns in multi-column models will significantly increase the size of the model, several works introduce special multi-path or multi-column modules cascaded into a single-column network. The first crowd counting model using this type of architecture was multi-scale CNN (MSCNN) \cite{MSCNN_ICIP2017}. MSCNN is a single-column CNN network that contains three multi-scale blobs (MSBs). Inspired from the Inception model \cite{Inception_CVPR2015}, the MSB is a naive Inception module (Fig. \ref{fig:MSB_MSCNN}) with four different sizes of Conv filters (i.e., $3\times3$, $5\times5$, $7\times7$, $9\times9$).

\begin{figure}[!h] \centering
\includegraphics[width=0.95\columnwidth]{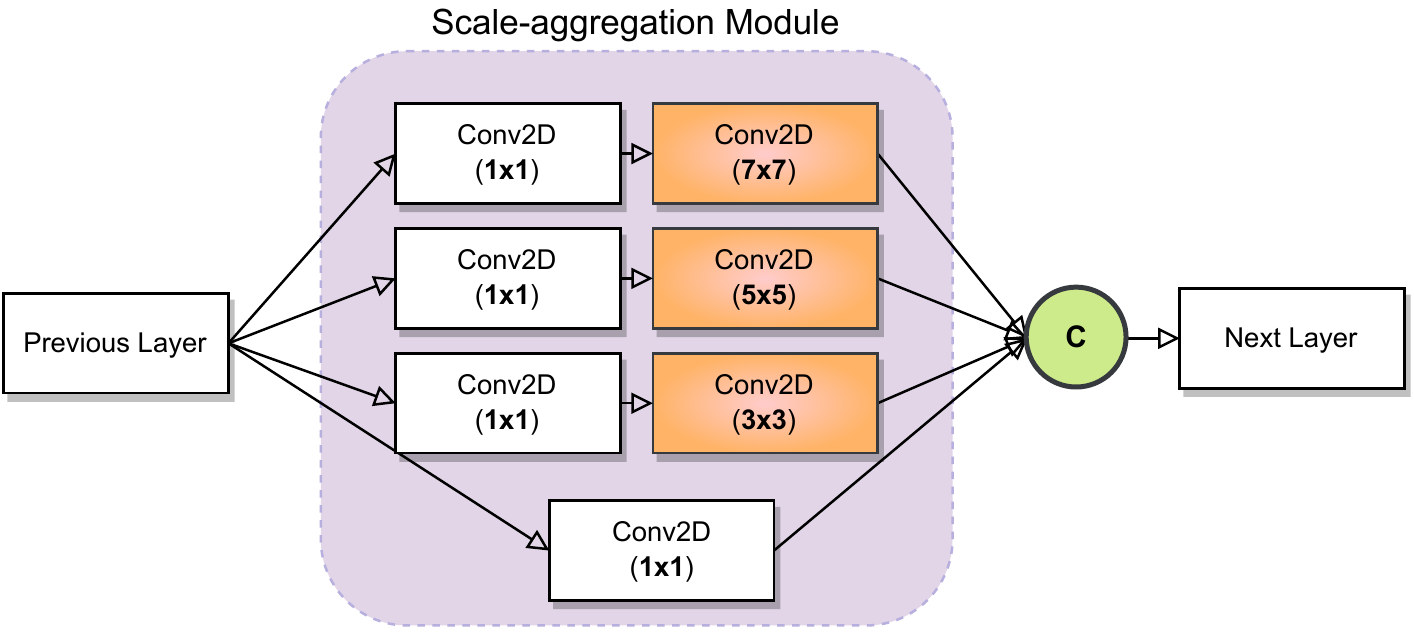}
\caption{Inception \cite{Inception_CVPR2015} like scale aggregation modules proposed in SANet \cite{SANet_ECCV2018}} \label{fig:SAM_SANet}
\end{figure}

A slightly different multi-scale module called as scale-aggregation module (Fig. \ref{fig:SAM_SANet}) is proposed in \cite{SANet_ECCV2018} using $1\times1$ convolution layers. Although not explicitly stated by authors, the scale-aggregate module is also an Inception-like module using $1\times1$ convolution. The $1\times1$ convolution layer when used before Conv layers of larger receptive field such as $5\times5$ and $7\times7$ apply dimension reduction and thus reduce computations.

Other examples include M-SegNet \cite{MSFANet-ICPR2021}, SGANet \cite{SGANet_IEEEITS2022}.

\subsection{Models with Transfer Learning}
Small crowd counting models typically suffer from accuracy degradation when used on high density scenes. To improve the accuracy of crowd models on highly dense scenes, transfer learning is becoming a promising approach. Although earlier models such as CrowdNet \cite{CrowdNet_CVPR2016} has been using this approach, its performance was limited. The first model that showed significant boost in performance using transfer-learning is congested scene recognition (CSRNet) \cite{CSRNet_CVPR2018}. CSRNet uses the first 10 layers of VGG-16 model \cite{VGG16_ICLR2015} pre-trained on ImageNet dataset \cite{imagenet_dataset2009} as front-end CNN. The back-end CNN used dilated-convolution instead of pooling operation making CSRNet an all-convolution network with larger receptive fields which is easier to train. Transfer-learning is typically used in any type of architecture i.e., single-column, multi-column, encoder-decoder etc.

\subsection{Other CNN Models}
Although, most crowd counting models inspires from other works and follow similar design patterns, there are few works that do not fit into either of the aforementioned categories. In other word, there is no particular design pattern followed and the network is formed by connecting various layers in different ways (i.e., cascaded, parallel, feedback paths, skip connections etc.). One such model is context-aware network (CANNet) \cite{CANNet_CVPR2019} which comprised of a VGG16 front-end (10 layers), proceeded by a multi-path network (consisting of convolution and up-sampling layers), which concatenates to connect to a single-column decoder network. Similarly, the perspective-guided
convolution network (PGCNet) in \cite{PGCNet_ICCV2019} uses a two column architecture. The first column is a density map prediction network (DMPNet), whereas the second column is the perspective estimation network (PENet). The DMPNet uses CSRNet \cite{CSRNet_CVPR2018} as a backbone, and then use multiple perspective-guided convolution (PGC) modules cascaded in a single column. The PENet on the other hand, uses an encoder-decoder CNN network to estimate the perspective map of the input image.

In \cite{MMCNN_ACCV2020}, authors propose multi-modal counting network that accepts both RGB images and infrared images of the same scene as inputs to predict the density maps. The MMCNN architecture is also a complex network with inter-connections among various layers at different stages. MMCNN also uses ResNet \cite{ResNet_CVPR2016} based multi-scale modules.

\subsection{Transformer-based Models}
Transfomer models proposed in \cite{transformers_NIPS2017} have recently emerged as a new paradigm of neural networks with self-attention mechanism. Due to their remarkable performance in several applications such as natural language processing \cite{} and speech processing \cite{}. These profound performance of transfomers in other domains has sparked the computer vision community to apply transfomers in various vision applications. Vision transformer (ViT) \cite{ViT_ICLR2021} is the first transformer model applied in computer vision for image classification. Since then, transformers have been used in various vision applications including image recognition \cite{Touvron_2021}, object detection \cite{Carion_2020,Zhu_2021}, and image segmentation \cite{Ye_2019}. 

The original transformer \cite{transformers_NIPS2017} uses an encoder-decoder structure with self-attention mechanism. The encoder transform an input sequence (words in NLP) into a high dimensional feature vector whereas the decoder then translate it into the output sequence. When applied in vision tasks \cite{ViT_ICLR2021}, the input image is split into patches and then the patches are provided to the transformer's encoder as linear embeddings (to mimic sequences).
Recently, transformers have been applied in crowd counting tasks \cite{Ranjan_2019, DCST_2021, CCTrans_2021, TAM_2021, Do_2021, Liang_2022, Wei_2021, TransCrowd_2022}. For example, the TransCrowd model \cite{TransCrowd_2022} uses pure transformer model in crowd counting task. It convert each image of size (1152x768) into fixed size six (6) patches of size (384x384) and convert patches to sequences of pixel values in the patch (flattening). The sequences are fed to the encoder of the ViT transformer (pretrained on ImageNet) \cite{ViT_ICLR2021}. The output is then fed to a regression head which predict the total count in the image. Unlink TransCrowd \cite{TransCrowd_2022} which predicts the total count, authors in \cite{CCTrans_2021} propose CCTrans i.e., a transformer model to predict density maps. CCTrans uses similar input pipeline as TransCrowd but uses different model architecture. It uses Twins \cite{Twins_2021} as backbone as feature encoder. The 1D output is transformed into feature maps (2D vectors), upsampled to $\frac{1}{8}$ of the input patch size. The feature maps are then fed to regression heads to predict density maps.

Some recent works also propose alternate techniques for crowd counting such as deep q-learning (DQN) \cite{Liu_2020, Xu_2021}. However, such contributions have limited impact in terms of performance improvement.


\begin{figure*}
\centering 
\subcaptionbox{Single-column models depict convolution layers cascaded in a single column.}{\includegraphics[align=c, width=0.45\textwidth]{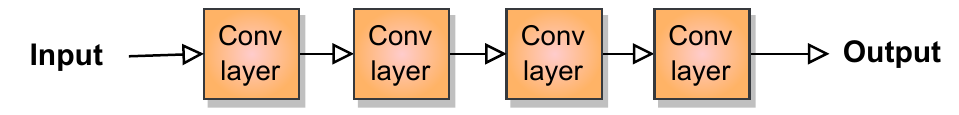}} \hspace{2em}
\subcaptionbox{Multi-column models consist of two or more CNN layers where the output of all columns are fused together to produce a single output.}{\includegraphics[align=c, width=0.45\textwidth]{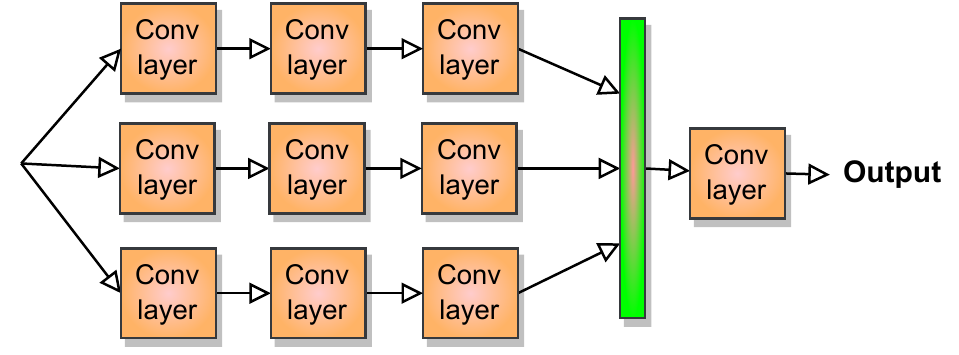}}
\\ \vspace{3em}

\subcaptionbox{Encoder-decoder models comprised of two parts, where the first part (encoder module) of the model works as a feature extractor whereas the second part (decoder module) of the network decodes the features into the required output shape.}{\includegraphics[width=0.45\textwidth]{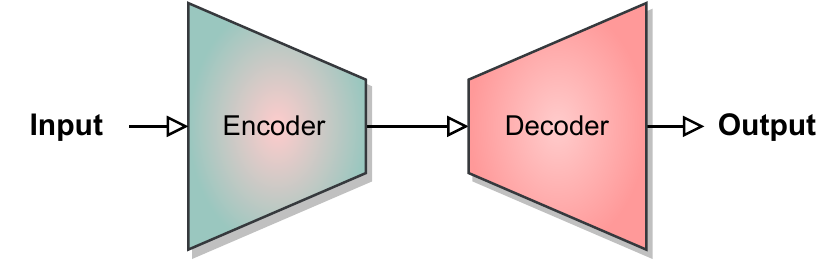}} \hspace{2em}
\subcaptionbox{Models which consist of a relatively smaller CNN network with a pre-trained backbone or front-end network (large sized).}{\includegraphics[width=0.45\textwidth]{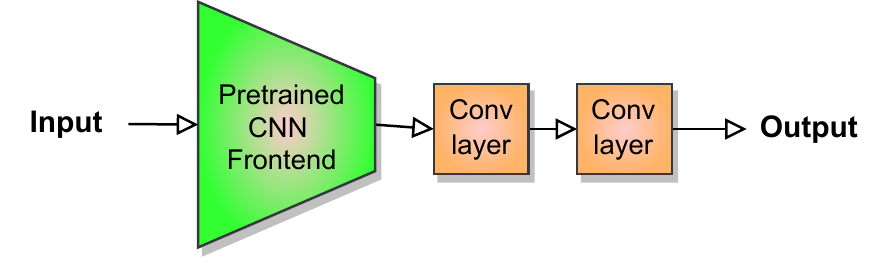}}
\\ \vspace{3em}

\subcaptionbox{A single column model with multiple modules which are usually multi-column, specifically designed to encode scale-aware feature in crowd images.}{\includegraphics[width=0.45\textwidth]{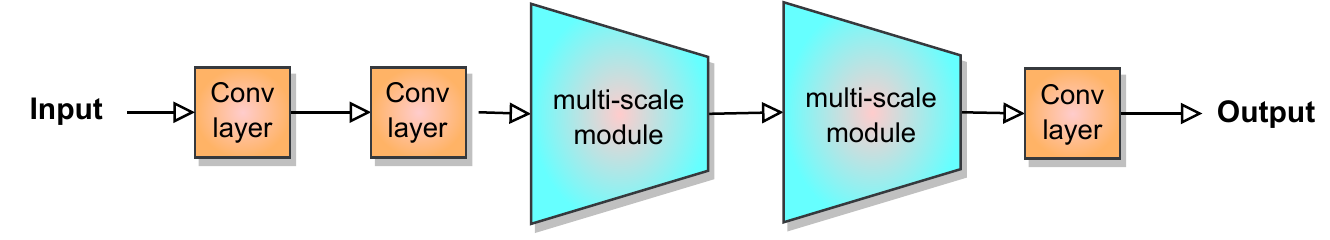}} \hspace{2em}
\subcaptionbox{Transfomer Models.}{\includegraphics[width=0.45\textwidth]{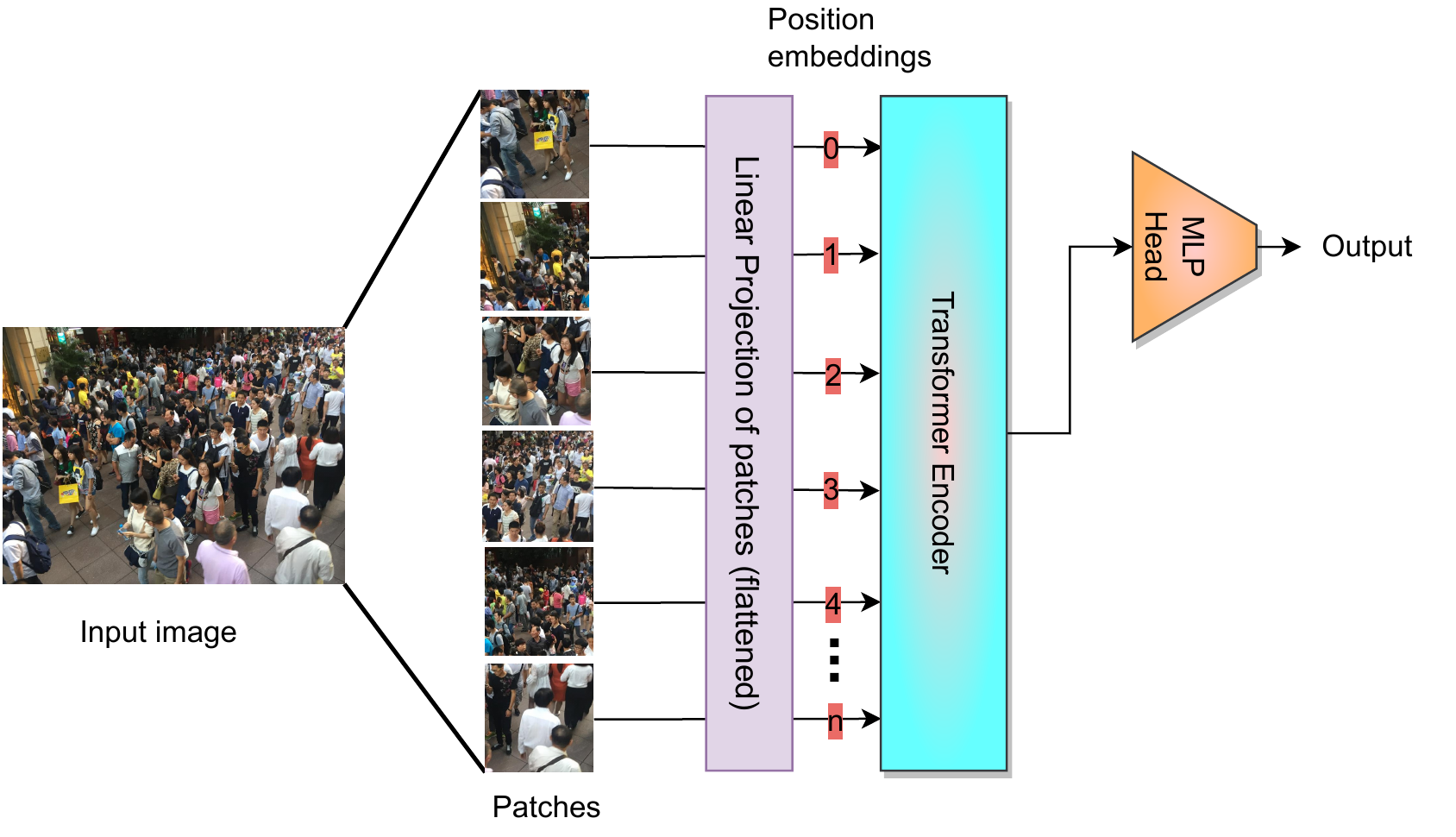}}

\caption{Crowd counting models: Common CNN design patterns followed in crowd counting and density estimation models.}
\label{fig:cc_nets}
\end{figure*}

\subsection*{Summary and Lessons Learned}
\begin{itemize}
\item Single-column models are compact but achieves lower accuracy over dense images. They also suffer from scale variations in images.
\item Scale variations can be addressed using multi-column models. However, the capability of a multi-column model to adopt to various object scales is limited by the number of columns. Multi-scale models are also computationally expensive due to several columns training in parallel.
\item To address scale variations in a computationally efficient way and to adopt to large scale variations, single column models with multi-scale modules are proposed. These works are originally inspired from Inception \cite{Inception_CVPR2015} model with small differences in design.
\item Encoder-decoder models preserve spatial resolution when high quality density maps are desired. They also offer easy implementation of supervision at different stages of the network (e.g., TEDnet \cite{TEDnet_CVPR2019}).
\item Both single-column and multi-column models trained from scratch have limited accuracy when tested on large datasets with highly dense images. Hence, pre-trained deep models such as VGG \cite{VGG16_ICLR2015}, Inception \cite{Inception_CVPR2015}, and ResNets \cite{ResNet_CVPR2016} have been used to improve counting accuracy. Models with pre-trained backbone CNNs (front-ends) also train faster. However, the model size and execution time increases which make such models not a good choice for real-time performance.
\end{itemize}


\begin{table*}[]
\caption{A list of crowd counting models arranged in chronological order.} \label{tab:counting_sota}
\centering
\small
\begin{adjustbox}{max width=0.9\textwidth}
\begin{tabular}{r|c|c|c| cccccccccccccc} \toprule
& & &
& \multicolumn{14}{c} {Datasets} \\ \cmidrule{5-18}

Model &Year  &Architecture &Backbone
&\rot{Mall \cite{Mall_dataset2012}} 
&\rot{UCSD \cite{UCSD_dataset2008}} 
&\rot{WorldExpo'10 \cite{CrowdCNN_CVPR2015}} 
&\rot{UCF-CC-50 \cite{UCF_CC_50_dataset2013}} 
&\rot{ShanghaiTech \cite{MCNN_CVPR2016}} 
&\rot{UCF-QNRF \cite{CompositionLoss_2018}}
&\rot{SmartCity \cite{SaCNN_WACV2018}} 
&\rot{DroneRGBT \cite{SaCNN_WACV2018}} 
&\rot{CARPK \cite{SaCNN_WACV2018}} 
&\rot{TRANCOS \cite{TRANCOS_dataset2015}}
&\rot{VisDrone \cite{SaCNN_WACV2018}}
&\rot{RGBT-CC \cite{RGBTCC_dataset2021}}
&\rot{JHU-Crowd \cite{jhucrowd_dataset2020}}
&\rot{NWPU-Crowd \cite{NWPUCrowd_dataset2021}}
\\ \midrule \midrule

CrowdCNN \cite{CrowdCNN_CVPR2015} &2015 
& SC
& No
&\xmark &\cmark &\cmark &\cmark &\xmark &\xmark &\xmark &\xmark &\xmark &\xmark &\xmark &\xmark &\xmark  &\xmark\\[1.5em]


MCNN \cite{MCNN_CVPR2016} &2016 
& MC
& No
&\xmark &\cmark &\cmark &\cmark &\cmark &\xmark &\xmark &\xmark &\xmark &\xmark &\xmark &\xmark &\xmark &\xmark\\[0.2em]

CrowdNet \cite{CrowdNet_CVPR2016} &2016 
& MC
& VGG16
&\xmark &\xmark &\xmark &\cmark &\xmark &\xmark &\xmark &\xmark &\xmark &\xmark &\xmark &\xmark &\xmark &\xmark\\[1.5em]


MSCNN \cite{MSCNN_ICIP2017} &2017 
& SC-MSM
& No
&\xmark &\xmark &\xmark &\cmark &\cmark &\xmark &\xmark &\xmark &\xmark &\xmark &\xmark &\xmark &\xmark &\xmark\\[0.2em]

SCNN \cite{SCNN_CVPR2017} &2017 
& MC
& No
&\xmark &\cmark &\cmark &\xmark &\cmark &\xmark &\xmark &\xmark &\xmark &\xmark &\xmark &\xmark &\xmark &\xmark\\[0.2em]

CMTL \cite{CMTL_AVSS2017} &2017 
& MC
& No
&\xmark &\xmark &\xmark &\cmark &\cmark &\xmark &\xmark &\xmark &\xmark &\xmark &\xmark &\xmark &\xmark &\xmark\\[1.5em]


CSRNet \cite{CSRNet_CVPR2018} &2018 
& SC
& VGG16
&\xmark &\cmark &\cmark &\cmark &\cmark &\xmark &\xmark &\xmark &\xmark &\xmark &\xmark &\xmark &\xmark &\xmark\\[0.2em]

SANet \cite{SANet_ECCV2018} &2018 
& ED/SC-MSM
& No
&\xmark &\cmark &\cmark &\cmark &\cmark &\xmark &\xmark &\xmark &\xmark &\xmark &\xmark &\xmark &\xmark &\xmark\\[0.2em]

SaCNN \cite{SaCNN_WACV2018} &2018 
& SC
& No
&\xmark &\xmark &\cmark &\cmark &\cmark &\xmark &\cmark &\xmark &\xmark &\xmark &\xmark &\xmark &\xmark &\xmark \\[0.2em]

DecideNet \cite{DecideNet_CVPR2018} &2018 
& MC
& No
&\xmark &\xmark &\cmark &\xmark &\cmark &\xmark &\xmark &\xmark &\xmark &\xmark &\xmark &\xmark &\xmark &\xmark \\[1.5em] 


TEDnet \cite{TEDnet_CVPR2019} &2019 
& ED
& No
&\xmark &\xmark &\cmark &\cmark &\cmark &\cmark &\xmark &\xmark &\xmark &\xmark &\xmark &\xmark &\xmark &\xmark \\[0.2em]

CANNet \cite{CANNet_CVPR2019} &2019 
& Complex
& VGG16
&\xmark &\xmark &\cmark &\cmark &\cmark &\cmark &\xmark &\xmark &\xmark &\xmark &\xmark &\xmark &\xmark &\xmark \\[0.2em]  

GSP \cite{GSP_CVPR2019} &2019 
& SC
& VGG16
&\xmark &\xmark &\xmark &\xmark &\cmark &\xmark &\xmark &\xmark &\xmark &\xmark &\xmark &\xmark &\xmark &\xmark \\[0.2em] 

MobileCount \cite{MobileCount_PRCV2019} &2019 
& ED
& MobileNetV2
&\xmark &\xmark &\cmark &\cmark &\cmark &\cmark &\xmark &\xmark &\xmark &\xmark &\xmark &\xmark &\xmark &\xmark \\[0.2em] 

ANF \cite{ANF_CVPR2019} &2019 
& ED
& No
&\xmark &\xmark &\cmark &\cmark &\cmark &\cmark &\xmark &\xmark &\xmark &\xmark &\xmark &\xmark &\xmark &\xmark \\[0.2em]

BL \cite{BL_ICCV2019} &2019 
& SC
& VGG19
&\xmark &\xmark &\cmark &\cmark &\cmark &\cmark &\xmark &\xmark &\xmark &\xmark &\xmark &\xmark &\xmark &\xmark \\[0.2em]

PACNN \cite{PACNN_CVPR2019} &2019
& MC
& VGG16
&\xmark &\cmark &\cmark &\cmark &\cmark &\xmark &\xmark &\xmark &\xmark &\xmark &\xmark &\xmark &\xmark &\xmark \\[0.2em]

PGCNet \cite{PGCNet_ICCV2019} &2019
& Complex
& CSRNet
&\xmark &\xmark &\cmark &\cmark &\cmark &\xmark &\xmark &\xmark &\xmark &\xmark &\xmark &\xmark &\xmark &\xmark \\[1.5em]



C-CNN \cite{CCNN_ICASSP2020} &2020 
& SC
& VGG
&\xmark &\xmark &\cmark &\xmark &\cmark &\xmark &\xmark &\xmark &\xmark &\xmark &\xmark &\xmark &\xmark &\xmark \\[0.2em]

DeepCount \cite{DeepCount_ECAI2020} &2020 
& MC
& CSRNet
&\xmark &\xmark &\xmark &\xmark &\cmark &\cmark &\xmark &\xmark &\xmark &\xmark &\xmark &\xmark &\xmark &\xmark \\[0.2em] 

ASNet \cite{ASNet_CVPR2020} &2020 
& MC
& VGG16
&\xmark &\xmark &\cmark &\cmark &\cmark &\cmark &\xmark &\xmark &\xmark &\xmark &\xmark &\xmark &\xmark &\xmark \\[0.2em]

MMCNN \cite{MMCNN_ACCV2020} &2020
& Complex
& ResNet-50
&\xmark &\xmark &\xmark &\xmark &\xmark &\xmark &\xmark &\cmark &\xmark &\xmark &\xmark &\xmark &\xmark &\xmark \\[0.2em]

LibraNet \cite{LibraNet_ECCV2020} &2020
& SC
& VGG16
&\xmark &\xmark &\xmark &\cmark &\cmark &\cmark &\xmark &\xmark &\xmark &\xmark &\xmark &\xmark &\xmark &\xmark \\[0.2em]

MATT \cite{MATT_Elsevier_PR2020} &2020
& MC
& CSRNet
&\xmark &\xmark &\xmark &\xmark &\xmark &\xmark &\xmark &\xmark &\xmark &\xmark &\xmark &\xmark &\xmark &\xmark \\[0.2em] 


Yang et al. \cite{Yang_ICCV2020} &2020
& MC
& VGG16
&\xmark &\cmark &\cmark &\xmark &\cmark &\xmark &\xmark &\xmark &\xmark &\xmark &\xmark &\xmark &\xmark &\xmark \\[0.2em]

PCCNet \cite{PCCNet_IEEE_TCSVT2020} &2020
& MC
& No
&\xmark &\xmark &\cmark &\cmark &\cmark &\cmark &\xmark &\xmark &\xmark &\xmark &\xmark &\xmark &\xmark &\xmark
\\[1.5em]


SASNet \cite{SASNet_AAAI2021} &2021 
& ED
& VGG16
&\xmark &\cmark &\cmark &\cmark &\cmark &\xmark &\xmark &\xmark &\xmark &\xmark &\xmark &\xmark &\xmark &\xmark \\[0.2em]

M-SFANet \cite{MSFANet-ICPR2021} &2021  
& SC-MSM
& VGG16
&\xmark &\xmark &\cmark &\cmark &\cmark &\cmark &\xmark &\xmark &\xmark &\xmark &\xmark &\xmark &\xmark &\xmark \\[0.2em] 

M-SegNet \cite{MSFANet-ICPR2021} &2021  
& SC-MSM
& VGG16
&\xmark &\xmark &\cmark &\cmark &\cmark &\cmark &\xmark &\xmark &\xmark &\xmark &\cmark &\xmark &\xmark &\xmark \\[1.5em] 


SGANet \cite{SGANet_IEEEITS2022} &2022 
& SC-MSM 
& Inception-v3
&\xmark &\xmark &\xmark &\cmark &\cmark &\cmark &\xmark &\xmark &\xmark &\xmark &\xmark &\xmark &\xmark &\xmark \\[0.2em]

FusionCount \cite{FusionCount_2022} &2022
& SC
& VGG16
&\xmark &\xmark &\xmark &\xmark &\cmark &\xmark &\xmark &\xmark &\xmark &\xmark &\xmark &\xmark &\xmark &\xmark
\\[0.2em]

MAN \cite{MAN_CVPR2022} &2022
& MC
& VGG19
&\xmark &\xmark &\xmark &\xmark &\xmark &\cmark &\xmark &\xmark &\xmark &\xmark &\xmark &\xmark &\cmark &\cmark
\\[0.2em]

MFCC \cite{MFCC_2022} &2022
& ED
& ResNet
&\xmark &\xmark &\xmark &\xmark &\xmark &\xmark &\xmark &\cmark &\xmark &\xmark &\xmark &\xmark &\xmark &\xmark
\\[0.2em]

TAFNet \cite{TAFNET_2022} &2022
& MC
& VGG16
&\xmark &\xmark &\xmark &\xmark &\xmark &\xmark &\xmark &\xmark &\xmark &\xmark &\xmark &\cmark &\xmark &\xmark 
\\[0.2em]

TransCrowd \cite{TransCrowd_2022} &2022
& MC
& Non-CNN
&\xmark &\cmark &\cmark &\cmark &\cmark &\xmark &\xmark &\xmark &\xmark &\xmark &\xmark &\xmark &\cmark &\cmark
\\[0.2em] 
\bottomrule
\end{tabular}
\end{adjustbox}
\end{table*}

\section{Network Training and Evaluations} \label{sec:train_eval}
The standard process for supervised learning using neural network comprised of two stages: network training, and evaluation. In training stage, the network takes images and ground truths (total head count or density maps), compute gradients and back propagate the loss. The loss function is the objective function that the network is minimizing. After each one or every few epochs, the model is evaluated using a metric function. In the case of crowd counting, both the loss and metric functions compare the predicted density map with the ground truth (i.e., target) density map and compute the difference. The target and predicted density maps can be generally compared in three possible ways; (i) image-wise i.e., compare density maps as a whole, (ii) patch-wise i.e., divide the target and predicted maps into non-overlapping patches and compare all patches that belong to the same location, and (iii) pixel-wise i.e., compare each individual pixels (at the same location) of the target and predicted maps and integrate over the image.

\begin{figure}
\centering
\includegraphics[width=0.9\columnwidth]{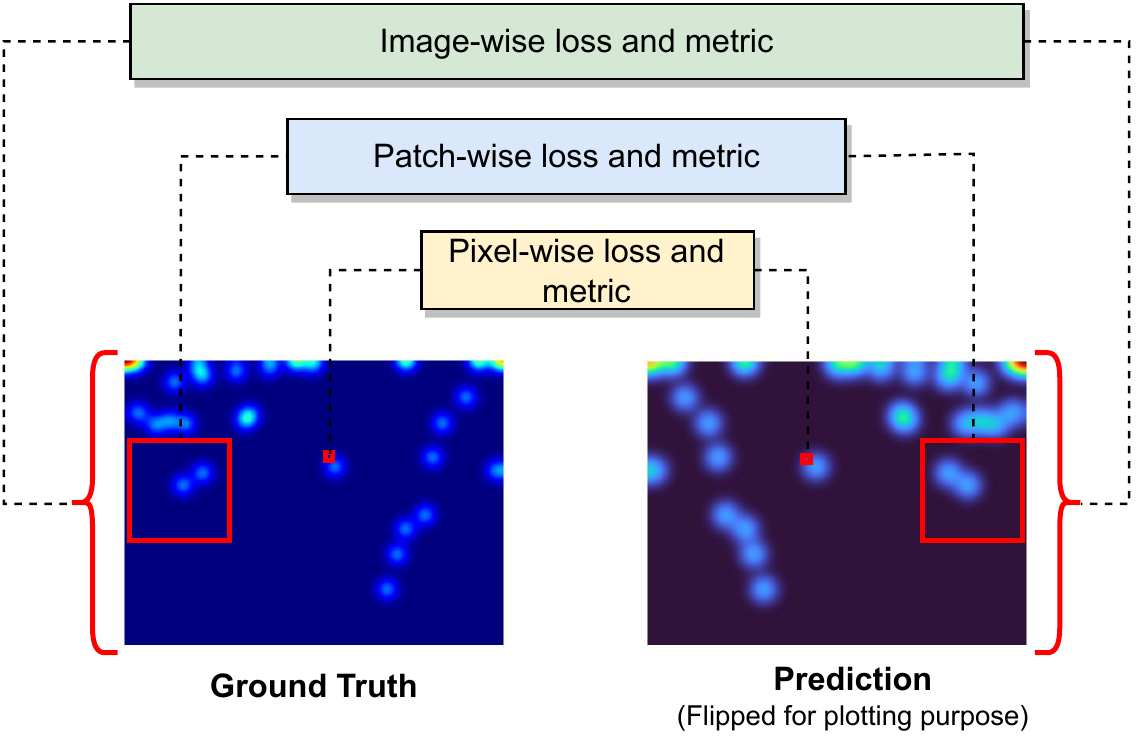}
\caption{Different levels of comparing predicted density maps with the target density maps.}
\label{fig:cc_learning}
\end{figure}

\subsection{Supervised and Weakly-supervised Learning}
The state-of-the-art methods for crowd density estimation use point-level supervision (fully supervised).
Although, fully supervised learning produces more reliable and accurate results, they require expensive annotations.

A small number of research works \cite{MATT_Elsevier_PR2020, Wang2019, TransCrowd_2022, Liu2019} also propose the use of weakly supervised learning which use more economical total head count annotations.
The Multiple Auxiliary Tasks Training (MATT) model \cite{MATT_Elsevier_PR2020} uses dot annotations (full supervision) for a small number of samples and count annotations (weak supervision) for the remaining data. According to the \cite{MATT_Elsevier_PR2020}, the count-level supervision although may achieve low counting errors, it does not encode precise object location. To overcome this, multiple auxiliary branches were used to produce density maps. The losses for these auxiliary branches are computed and integrated. The final loss function is the sum of pixel-wise $l_2$ loss (primary branch), count loss (primary branch) and the integrated auxiliary loss (auxiliary branches).

Recently transformer models are proposed to implement weakly supervised learning due to their self-attention mechanism to automatically learn the semantic information in crowd images. The TransCrowd model \cite{TransCrowd_2022} is based on the ViT \cite{ViT_ICLR2021} which uses only count level supervision to predict the count.

\subsection{Data Augmentation}
Data augmentation refers to the technique used to increase the amount of data and introduce more diversity in the data by adding slightly modified copies of original data. When the input data is images, different augmentation techniques are applied to transform original images such as random flipping, croping, resizing, random perspective, rotation, random noise, brightness, contrast etc. However, the augmentation techniques should be applied carefully depending on the problem. In crowd counting research, the most commonly applied augmentation techniques include random cropping \cite{CrowdCNN_CVPR2015, CrowdNet_CVPR2016, MCNN_CVPR2016, CSRNet_CVPR2018, DecideNet_CVPR2018, MSFANet-ICPR2021, PCCNet_IEEE_TCSVT2020} and horizontal flipping \cite{CSRNet_CVPR2018, DecideNet_CVPR2018, MSFANet-ICPR2021, PCCNet_IEEE_TCSVT2020}. In general, cropping helps to improve learning by addressing the scale variations whereas horizontal flipping address the perspective distortion in images \cite{CrowdNet_CVPR2016}. Some less-frequently used transformations include random noise \cite{DecideNet_CVPR2018}, vertical flipping \cite{DecideNet_CVPR2018}, and image rotation \cite{MMCNN_ACCV2020}. Image augmentation also acts as a regularize for the network i.e., helps to avoid overfit the model to the training data \cite{MMCNN_ACCV2020}.


\subsection{Loss functions}
A loss function is a function that measure the difference between the ground truth (GT) and the prediction. In density estimation, both the GT and prediction are density maps, whereas in count regression, both are numbers (integers).
\\

\subsubsection{Euclidean ($l_2$) loss:}
Euclidean loss or $l_2$ loss is the most common loss function used in crowd counting research. The pixel-wise $l_2$ loss or mean squared error($L_D(\Theta)$) is calculated as follows:

\begin{equation} \label{eq:loss_l2_pixelwise}
    L_D(\Theta) = \frac{1}{N} \sum_{i=1}^{N}{||F(X_i;\Theta) - D_i||^2}
\end{equation}
where $\Theta$ represents the CNN parameters (i.e., wights), $N$ is the total number of samples, $F(X_i; \Theta)$ is the predicted density map, and $D_i$ is the ground truth density map.

The $l_2$ loss can also be calculated over total counts instead of pixel-wise values. If 

\begin{equation} \label{eq:loss_l2_count}
    L_C(\Theta) = \frac{1}{N}{\sum_{i}^{N}{||\hat{C}_i - C_i ||^2}}
\end{equation}
where $\hat{C}_i$ and $C_i$ are the predicted and true counts, which are obtained by summing the pixel values over entire GT density map and predicted density map respectively.
\\

\subsubsection{$l_1$ loss} This is the Manhattan distance ($l_1$ norm) among corresponding pixels in the target and predicted density maps.

\begin{equation} \label{eq:loss_l1_pixelwise}
    L_D(\Theta) = \frac{1}{N} \sum_{i=1}^{N}{||F(X_i;\Theta) - D_i||}
\end{equation}

Similar to the case of $l_2$ loss, $l_1$ loss can also be computed over total count instead of individual pixels as follows:

\begin{equation} \label{eq:loss_l1_count}
    L_C(\Theta) = \frac{1}{N}{\sum_{i}^{N}{|| \hat{C}_i - C_i ||}}
\end{equation}

\subsubsection{Composition Loss} Composition loss is proposed in \cite{CompositionLoss_2018}. Two density maps and one localization map is generated at different depths in the CNN network, whereas the final output is the head count in the image. For the density and localization maps, the pixel-wise $l_2$ loss is computed, whereas the regression $l_2$ loss between the predicted and actual counts is computed.
\\

\subsubsection{Spatial Correlation Loss} Spatial correlation loss (SCL) is the cross-correlation among the normalized target and predicted density maps and is calculated as:
\begin{equation} \label{eq:loss_SC}
    L_{SC} =  1 - \frac{ \sum_{i=1}^{N} \sum_{j=1}^{M} \left(\hat{D_{ij}}. D_{ij} \right )}   {\sqrt{\sum_{i=1}^{N}{\hat{D}_{ij}^2}  . \sum_{i=1}^{N}{D_{ij}^2} }}
\end{equation}
where $D_{ij}$ and $\hat{D_{ij}}$ are the target and predicted density maps, respectively. $i$ and $j$ are the row and column indices, and $N$, and $M$ represents the total number of rows and columns in the target and predicted density maps, respectively. SCL has been used in TEDnet \cite{TEDnet_CVPR2019} in combination with $l_2$ loss. SCL is less-sensitive to changes in intensity levels in the density map.
\\
\subsubsection{APLoss}
The commonly used $l_2$ loss takes the target and predicted density maps as a whole to compute the pixel-wise euclidean $l_2$ distance among them. The standard $l_2$ loss does not consider the various density levels in the image and lead to generalization errors. To cope with this shortcoming of $l_2$ loss, adaptive pyramid loss (APLoss) is proposed in \cite{ASNet_CVPR2020}. To compute APLoss, the target density map is first divided into four non-overlapping patches. If the count in each patch is greater than a threshold $T$, the patch is further divided into four patches. This process is repeated until no patch is left with count greater than the threshold $T$. When this patching process is completed, the predicted density map is also divided into patches in the same pattern (the no. of patches should equal for both target and predicted density maps). Then the APLoss is calculated as follows:

\begin{equation} \label{eq:loss_APLOSS}
    APLoss = \frac{1}{N} \sum_{i=1}^{N} \sum_{j_1=1}^{4}{l_{R_j1}^{i}}
\end{equation}
where $N$ is the number of density maps, and $l_{R_j1}^{i}$ represents the patch-wise loss.
\\

\subsubsection{PRA Loss}
The APLoss function introduced in \cite{ASNet_CVPR2020} aim at reducing estimation errors in the more dense regions of the image. The Pyramid Region Loss (PRA Loss) \cite{SASNet_AAAI2021} is a natural extension of the APLoss with two improvements. The intuition behind PRA Loss is that in each crowd image, some regions and hence some pixels in those regions are harder for the model to learn from. These regions and pixels are typically overestimated which contribute to more counting errors. If these harder regions and more specifically the harder pixels in these regions are determined and assigned higher weights in the loss function, the network would be able to learn these pixels more consistently. PRA Loss solves this problem by extending the APLoss function with few improvements. Similar to APLoss, the predicted density map is divided into four patches and the patch with highest counting error (over-estimated patch) is selected. The patching process is repeated until the pixels with the highest errors are determined. 
PRA Loss is calculated as:

\begin{equation} \label{eq:loss_PRALoss}
l_{PRA} = || D^{est}_{p \in G}  - D^{GT}_{p \in G} ||^2 + \lambda || D^{est}_{p \in H}  - D^{GT}_{p \in G} ||^2
\end{equation}
where $p$ is the pixel in the predicted density map $D^{est}$, $G$ represents set of all pixels in $D^{est}$, and $H$ is the set of all hard pixels in $D^{est}$, and $\lambda$ is the weight term for the hard pixels.
\\

\subsubsection{Curriculum Loss}
In machine learning, curriculum learning is a trained strategy for neural networks proposed by \cite{curriculum_learning_ICML2009}, in which the model is fed with training samples in the order of increasing complexity. Curriculum learning was used in crowd counting by \cite{} such that each image is assigned a difficulty score. Inspired by the work in \cite{PSDDN_CVPR2019}, a curriculum loss function is proposed in \cite{SGANet_IEEEITS2022}. To design the curriculum loss, the difficulty level of all pixels in the density map are calculated. Following the intuition that dense crowd regions are harder to estimate, the pixels with higher values than a dynamic threshold are considered as difficult. To compute the difficulty level and the final loss function, a weight matrix $W$ of the same size as the density map is defined. 
\\

\begin{figure}
\centering
\includegraphics[width=0.9\columnwidth]{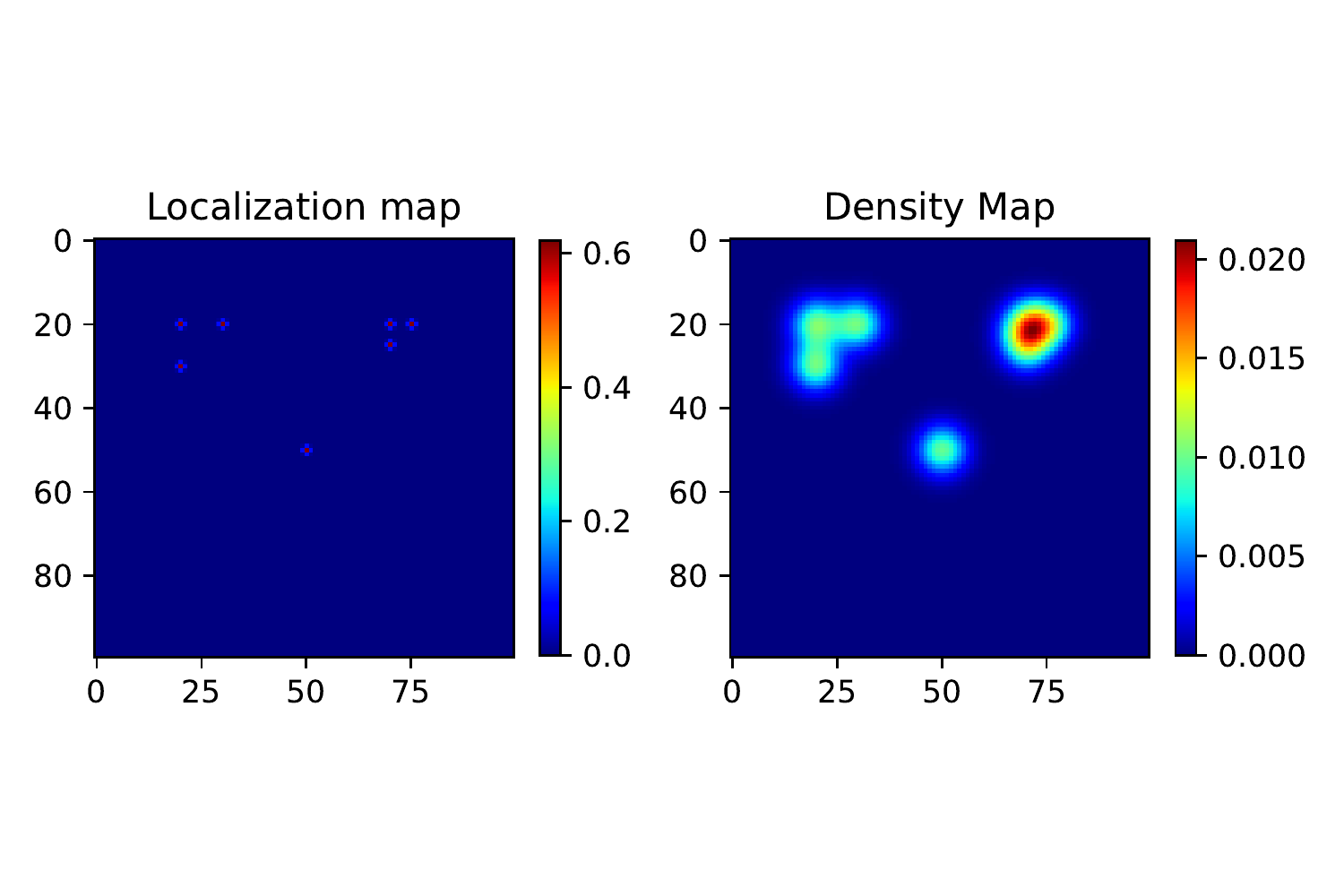}
\caption{An illustration of pixel values in density maps in different crowd density regions. The localization map shows location of heads (i.e., one head at the center, three heads at large distance in the top-left corner, and three heads closer to each other at top-right corner). The density map shows the corresponding pixel values i.e., higher values in the dense region (top-right).} \label{fig:pixels_density}
\end{figure}

\subsubsection{Combination Loss}
Several works propose a combination of two or more loss functions to improve learning. For instance, \cite{CrowdCNN_CVPR2015} proposes a combination of pixel-wise and count $l_2$ loss as follow:

\begin{equation} \label{eq:combination_loss_CrowdCNN}
    L(\Theta) =  L_D(\Theta) + \lambda L_Y(\Theta)
\end{equation}
where, $L_D(\Theta)$ is the pixel wise loss as calculated using Eq. \ref{eq:loss_l2_pixelwise}, and $\lambda L_Y(\Theta)$ is the $l_2$ loss over total count. The term $\lambda$ is the weight term.
\\

\subsubsection{Optimal Transport (OT) Loss}
The optimal transport (OT) loss is first introduced in \cite{DMCount_NeuroIPS2020}.
To compute the OT loss, both the target  and predicted density maps ($D$ and $\hat(D)$) are first converted into their respective probability density functions (pdfs) by dividing each pdf by its total mass. The OT loss is computed as follow:
\begin{equation} \label{eq:loss_OT}
    l_{OT} = \mathcal{W} \left( \frac{D}{\|D\|_1}, \frac{\hat{D}} {\|\hat{D}\|_1} \right)
\end{equation}

\subsubsection{Total Variation (TV) Loss}
Authors in \cite{DMCount_NeuroIPS2020} derived total variation (TV) loss due to the inherited time complexity for computing OT loss. It is defined as:

\begin{equation} \label{loss_TV}
    l_{TV} = \frac{1}{2} \left\|  \frac{D}{\|D\|_1}, \frac{\hat{D}} {\|\hat{D}\|_1}  \right\|_1
\end{equation}


\begin{table*}[]
\centering
\caption{Comparison of crowd counting models: model architecture, inputs and outputs} \label{tab:cc_sota2}
\small

\begin{adjustbox}{max width=0.9\textwidth}
\begin{tabular}{r|c|c|c|c|c}
\toprule
Model   &Input &Input size  &Output size &loss &Metric \\ \midrule \midrule

CrowdCNN \cite{CrowdCNN_CVPR2015}   &patch &$72\times72$  & $18\times18$ &combination $l_2$ loss &MSE, MAE \\[0.2em]


MCNN \cite{MCNN_CVPR2016}  &image &arbitrary  &input size/4 &$l_2$ loss &MSE, MAE \\[0.2em]

CrowdNet \cite{CrowdNet_CVPR2016}   &patch &$225\times225$  &input size/8 &$l_2$ loss &MSE, MAE \\[0.2em]


MSCNN \cite{MSCNN_ICIP2017} &patch &$225\times225$  &$6\times6$ &$l_2$ loss &MSE, MAE \\[0.2em]

SCNN \cite{SCNN_CVPR2017}  &patch &image size/3  &- &$l_2$ loss &MSE, MAE \\[0.2em]

CMTL \cite{CMTL_AVSS2017} &patch &image size/4  &input size &$l_2$ loss, cross-entropy &MSE, MAE \\[0.2em]


CSRNet \cite{CSRNet_CVPR2018}  &patch &image size/8  &input size/8 &$l_2$ loss &MSE, MAE, PSNR, SSIM \\[0.2em]

SANet \cite{SANet_ECCV2018}  &patch &image size/4  &input size &composite ($l_2 + l^{SSIM}$) loss &MSE, MAE \\[0.2em]

SaCNN \cite{SaCNN_WACV2018}  &patch &image size/8  & &$l_2$ loss &MSE, MAE \\[0.2em] 

DecideNet \cite{DecideNet_CVPR2018} &patch &$4x3$ &- &composite ($l_1 + l_2$) loss &MSE, MAE \\[0.2em]


TEDnet \cite{TEDnet_CVPR2019}  &image  &arbitrary  &- &SAL, SCL &MSE, MAE, PSIM, SSIM \\[0.2em]

CANNet \cite{CANNet_CVPR2019}  &patch  &$224\times224$  &- &$l_2$ loss &RMSE, MAE \\[0.2em] 

GSP \cite{GSP_CVPR2019}   &patch  &multiple  &- &$l_2$ loss &RMSE, MAE \\[0.2em] 

MobileCount \cite{MobileCount_PRCV2019} &patch &(image size x 0.8) &- &$l_2$ loss &MSE, MAE  \\[0.2em]

ANF \cite{ANF_CVPR2019} &image &arbitrary &- &$l_2$ loss &MSE, MAE  \\[0.2em] 

BL \cite{BL_ICCV2019} &patches &$256\times256$ &- &BL ($l^{Bayes}$)  &MSE, MAE \\[0.2em]

PACNN \cite{PACNN_CVPR2019} &patch &image size/4 &- &composite ($l_2 + l^{DSSIM}$) loss, &MSE, MAE \\[0.2em]

PGCNet \cite{PGCNet_ICCV2019} &image &arbitrary &- &$l_2$ loss &MSE, MAE \\[0.2em]



CCNN \cite{CCNN_ICASSP2020} &image &arbitrary  & &$l_2$ loss &MSE, MAE \\[0.2em]

DeepCount \cite{DeepCount_ECAI2020} &patch &$384\times512$  &multiple &weighted $l_1$ loss &RMSE, MAE \\[0.2em] 

ASNet \cite{ASNet_CVPR2020} &patch &$128\times128$  &- &APA loss &MSE, MAE \\[0.2em]

MMCNN \cite{MMCNN_ACCV2020} &image &- &- &composite loss &RMSE, MAE \\[0.2em]

LibraNet \cite{LibraNet_ECCV2020} &patch &image size/2 &- &$l_1$ loss &MSE, MAE \\[0.2em]

MATT \cite{MATT_Elsevier_PR2020} &image &arbitrary &- &composite ($l_1$ + $l_2$) loss &MSE, MAE, RER \\[0.2em]


Yang et al. \cite{Yang_ICCV2020} &- &- &- &composite ($l_2$ + cross-entropy) loss &MSE, MAE \\[0.2em]

PCCNet \cite{PCCNet_IEEE_TCSVT2020} &patches &$512\times680$ &- &$l_2$ loss &MSE, MAE\\[0.2em]


SASNet \cite{SASNet_AAAI2021} &patch &$128\times128$  &- &composite ($l_2 + l^{SSIM}$) loss &MSE, MAE \\[0.2em] 

M-SFANet \cite{MSFANet-ICPR2021} &image &arbitrary  &image size/2 &BL ($l^{Bayes}$) &RMSE, MAE, GAME \\[0.2em] 

M-SegNet \cite{MSFANet-ICPR2021} &image &arbitrary  &image size/2 &BL ($l^{Bayes}$) &RMSE, MAE, GAME \\[0.2em]


SGANet \cite{SGANet_IEEEITS2022} &patch &$128\times128$  &input size/4 &curriculum loss  &RMSE, MAE \\[0.2em]

FusionCount \cite{FusionCount_2022} &patch &$384\times512$ &- &$l_2$ loss &RMSE, MAE  \\[0.2em]

MAN \cite{MAN_CVPR2022} &patch &$512\times512$ &- &composite loss &MSE, MAE \\[0.2em]

MFCC \cite{MFCC_2022} &image &$512\times640$ &input size &$l_2$ loss &RMSE, MAE \\[0.2em]

TAFNet \cite{TAFNET_2022} &image &arbitrary &- &$l_2$ loss &RMSE, GAME \\[0.2em]

TransCrowd \cite{TransCrowd_2022} &image &$1152\times768$ &- &$l_1$ loss &MSE, MAE \\[0.2em]

\bottomrule
\end{tabular}
\end{adjustbox}
\end{table*}


\begin{figure*}[!h] 
\centering
\subcaptionbox{ShanghaiTech Part B \cite{MCNN_CVPR2016} and WorldExpo10 \cite{CrowdCNN_CVPR2015} datasets.}{\includegraphics[width=0.9\textwidth]{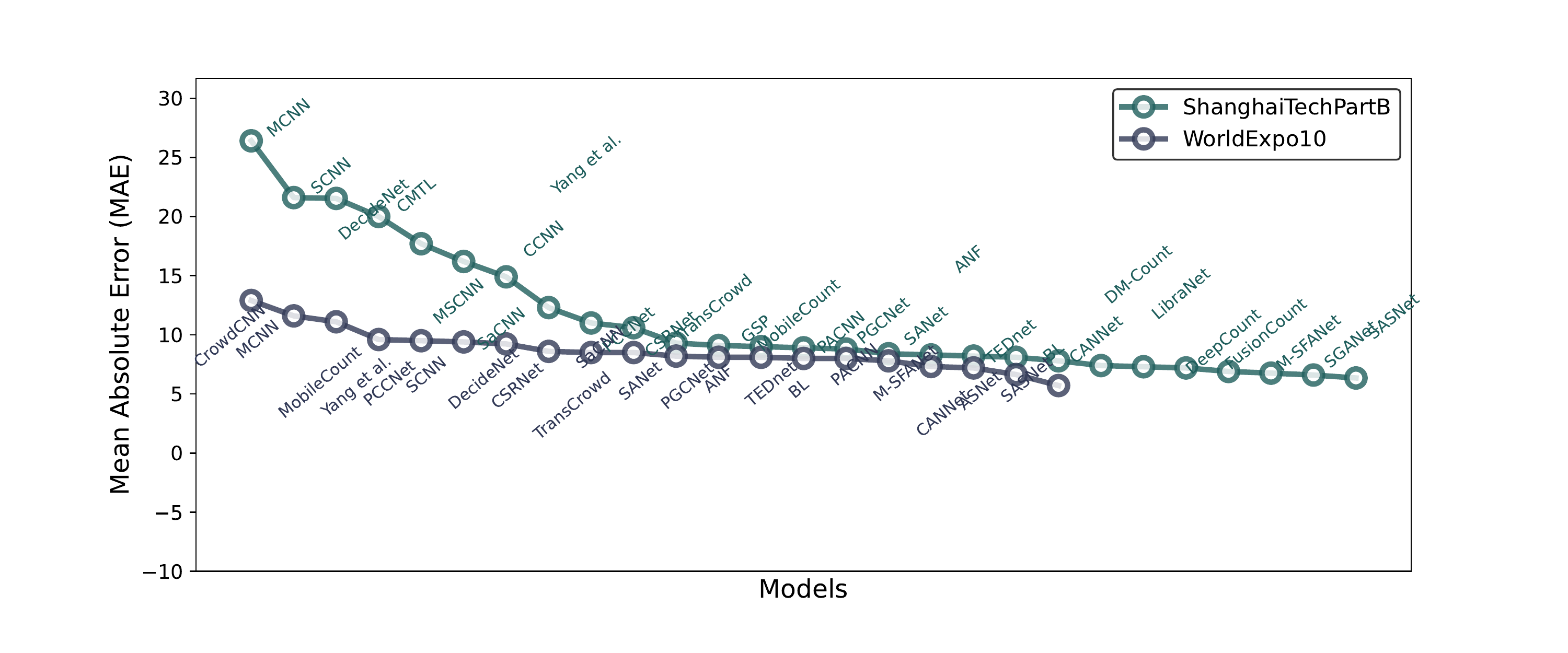}} \\
\vspace{0.5em}
\subcaptionbox{UCF-CC-50 \cite{UCF_CC_50_dataset2013}, ShanghaiTech Part A \cite{MCNN_CVPR2016} and UCF-QNRF \cite{CompositionLoss_2018} datasets.}{\includegraphics[width=0.9\textwidth]{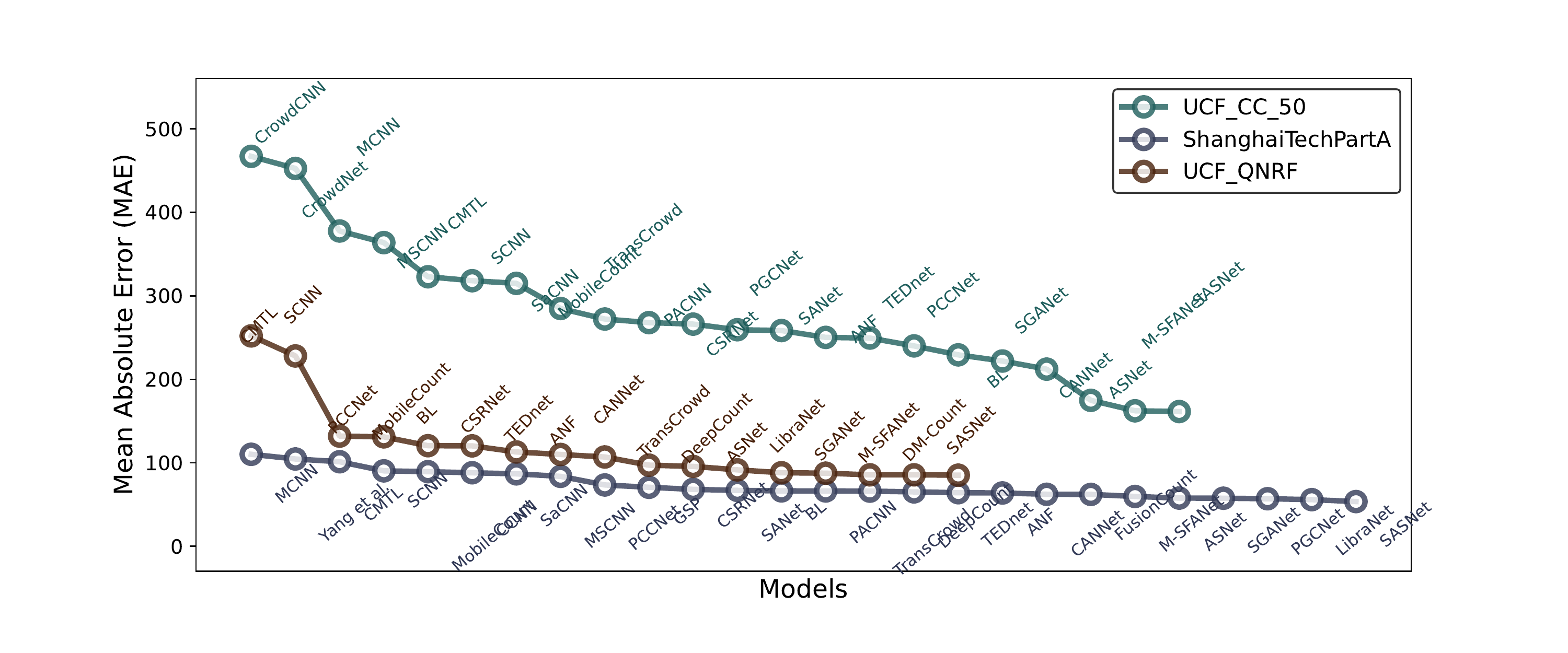}}
\caption{Performance analysis of crowd counting models over benchmark datasets.}
\label{fig:MAE_datasets}
\end{figure*}

\subsection{Evaluation Metrics}

\subsubsection{Mean Absolute Error (MAE)}
It is the $l_1$ or Manhattan distance between the actual and predicted count in an image and is calculated as follows:

\begin{equation}
    MAE = \frac{1}{N} \sum_{i=1}^{N}{(C^{pred}_i - \hat{C^{gt}_i})}
\end{equation}
MAE is the most commonly used metric in all crowd counting works. A limitation of MAE is that it is more robust to outliers (i.e., large counting errors). Thus mean squared error is used as an additional metric which is more sensitive to outliers.

\subsubsection{Mean Squared Error (MSE)}
\begin{equation}
    MSE = \frac{1}{N} \sum_{i=1}^{N}{(C^{pred}_i - \hat{C^{gt}_i})}^2
\end{equation}

where, $N$ is the total number of images in the dataset, $C^{gt}_i$ is the ground truth (actual count) and $C^{pred}_i$ is the prediction (estimated count) in the $i^{th}$ image. MSE is the second most used metric after MAE in crowd counting research.

\subsubsection{Grid Average Mean Error (GAME)}
While MAE is the most widely used metric in crowd counting research and is often used to compare various models, MAE provide image-wide counting and does not provide where the estimations have been done in the image. Owing to the possible estimation errors in MAE, authors in \cite{GAME_metric2015} proposed Grid Average Mean absolute Error (GAME). In GAME, an image is divided into $4^L$ non-overlapping patches and compute MAE separately for each patch. Thus, GAME poses a more robust and accurate estimation for crowd counting applications.

\begin{equation}
    GAME = \frac{1}{N} \sum_{i=1}^{N}{ ( \sum_{l=1}^{4^L}{|C^{pred}_{(n,l)} - C^{gt}_{(n,l)}|)}}
\end{equation}

\begin{figure}[!h]
\centering
\includegraphics[width=0.8\columnwidth]{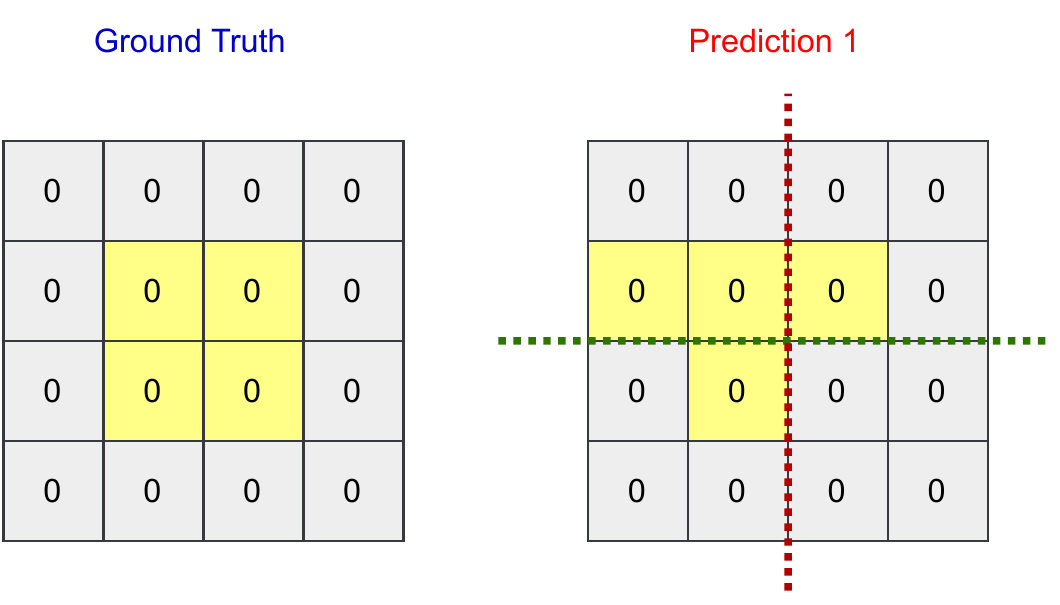}
\caption{An illustration of robustness of GAME metric to localization errors. For a single prediction of a given ground truth, MAE is calculated as zero (ignoring localization error in prediction), whereas GAME=1 (capturing localization error).} \label{fig:GAME}
\end{figure}

The GAME metric is more robust to localisation errors in density estimation. It can be easily observed in Fig. \ref{fig:GAME}.
\\

\subsubsection{Patch Mean Absolute Error (PMAE)}

\begin{equation} \label{eq:metric_PMAE}
    PMAE =  \frac{1}{m \times N} \sum_{i=1}^{m \times N}{|C^{pred}_{I_i} - C^{gt}_{I_i}|}
\end{equation}
where $m$ is the number of non-overlapping patches in each image. When $m=1$, PMAE become equivalent to MAE. PMAE is effectively equivalent to GAME for $m = 4^L$.
\\

\subsubsection{Patch Mean Squared Error (PMSE)}
\begin{equation} \label{eq:metric_PMSE}
    PMSE =  \frac{1}{m \times N} \sum_{i=1}^{m \times N}{(C^{pred}_{I_i} - C^{gt}_{I_i})^2}
\end{equation}
where $m$ is the number of non-overlapping patches in each image. When $m=1$, PMSE become equivalent to MSE.
\\

\subsubsection{Mean Pixel Level Absolute Error (MPAE)} This is equivalent to pixel-wise $l_1$ error and is calculated using Eq. \ref{eq:loss_l1_pixelwise}.

In addition to metrics for counting accuracy, some works \cite{CSRNet_CVPR2018, TEDnet_CVPR2019} used additional metrics to measure the quality of predicted density maps. These are listed as follows:
\\

\subsubsection{Peak Signal to Noise Ratio (PSNR)}
PSNR measures the error between the corresponding pixels in the target and predicted density maps. The value of PSNR is always in the range of [0-1]. Higher value of PSNR means lower error and vice versa. PSNR metric is used in \cite{CSRNet_CVPR2018, TEDnet_CVPR2019}.
\\

\subsubsection{Structural Similarity Index (SSIM)}
SSIM measures the difference between the target and predicted density maps over three attributes i.e., brightness, contrast, and structure. Similar to PSNR, the value of SSIM is also in the range of [0-1] and the higher value is preferred over lower values. SSIM metric is used in \cite{CSRNet_CVPR2018, TEDnet_CVPR2019}.
\par

A summary of different learning methods and evaluation criteria is presented in Table \ref{tab:cc_sota2}. We further compared the performance of these models over the commonly used metric i.e., MAE in Fig. \ref{fig:MAE_datasets} over benchmark datasets.

\subsection*{Summary and Lessons Learned}
\begin{itemize}
\item The pixel-wise euclidean or $l_2$ loss (in Eq. \ref{eq:loss_l2_pixelwise}) is the most commonly used loss function used almost in all crowd counting models. It is typically used alone or in combination with other loss functions to improve learning performance and convergence.

\item The euclidean loss can not detect the localization errors. Hence, more complex loss functions are proposed such as APLoss which attentively divide target density maps into patches or sub-regions and compute loss loss for each patch and finally sum these losses to calculate loss for the whole density map.

\item Almost all crowd counting models use $l_2$ distance among respective pixels in the target and predicted density maps. There is an underlying assumption that the individual pixel values are independent of their neighbors (non-correlated). Recently, authors in \cite{NoisyCC_NeuroIPS2020} show that how human annotation errors can cause noise in density maps generation and that such noise can be modeled as a multi-variate Gaussian approximation.

\item Despite the various loss functions proposed over time, the pixel-wise $l_2$ loss is the widely used function to compute loss in crowd density estimation either directly or integrated in other complex loss functions such as APLoss,  PRA Loss etc.
\end{itemize}

\section{Conclusion and Future Research Directions} \label{sec:conclusion}
The paper presents a comprehensive review of the most significant contributions in the area of crowd counting using computer vision and deep learning. Based on our review of the literature, we mark few interesting observations listed as follows. In addition, we present insights on each observation for prospective readers.

\begin{itemize}

\item \textit{\textbf{State-of-the-art: in brief?}}
Despite very deep models and complex architectures, the accuracy gains over dense and large datasets is reasonably low (e.g., ShanghaiTech Part A \cite{MCNN_CVPR2016}, UCF-QNRF \cite{CompositionLoss_2018}, JHU-Crowd \cite{jhucrowd_dataset2020} etc.). A large amount of research works contribute incrementally very low accuracy gains over these datasets. In very dense crowd counting, there is still sufficient room for accuracy gains which incites further investigation. Two recent research directions in this area are improved loss functions and novel architecture such as vision transformer-based models.

\item \textit{\textbf{Deep or shallow models: how to choose?}}
The attempts to gain higher accuracy over large datasets generally lead to deeper and more complex architectures, but on sparse datasets with low crowd density images (e.g., UCSD \cite{UCSD_dataset2008}, Mall \cite{Mall_dataset2012}, and ShanghaiTech Part B \cite{MCNN_CVPR2016}), shallow models provide reasonably sufficient accuracy and deeper models may not be required in scenarios captured in these datasets. Noteably, single-scene crowd counting is the simplest task, followed by sparse multi-scene crowd detection.

\item \textit{\textbf{Need for benchmarking:}}
Research efforts are mostly contributed towards achieving accuracy gains (even if very small gains), and the resulting model complexity is often ignored. This results in increasing model complexity for small and often negligible improvement in accuracy. We believe that benchmarking over model training and inference time of crowd models need to be investigated.

\item \textit{\textbf{Design patterns in crowd counting models:}}
While the initial crowd models followed a single column simple architecture, followed by other types such as multi-column, encoder-decoder etc, the most recent architectures are increasingly complex with no clear design patterns. Several works proposing novel architecture also adopted improved training method (e.g., new loss functions). The effective gain by the architecture and training method alone were not thoroughly investigated. Furthermore, the increasing complexity of crowd counting models makes it hard to understand the pattern which contributed in the accuracy and can preclude innovation and improvements in such models.

Based on our review and the aforementioned observations, we provide few useful insights for prospective researchers.

\item \textit{\textbf{Cross-scene counting and domain adaptation:}}
Most of the crowd counting models are trained and tested on the same benchmark datasets. In general the research on model generalization is still very limited. Few studies investigate models model generalization i.e., model pretrained on one dataset and then fine-tuned on other dataset. However, the results typically reported belongs to the fine-tuned model. This creates a huge gap on cross-scene model generalization of crowd counting models which needs further investigation. It would be interesting to see model generalization in a range of diverse scenarios e.g., CCTV images, drone images, indoor and outdoor scenarios. Furthermore, domain adaptations is another exciting direction in crowd counting research.

\item \textit{\textbf{Can we have a universal crowd counting model?}}
The potential applications of crowd counting will require the model to operate at various hardware platforms (e.g., servers, drones, cameras, mobile phones etc.) with varying computing capabilities. Also, the performance requirement vary by applications (e.g., real-time, non real-time), surveillance type (e.g., CCTV-based surveillance, drone-based surveillance) and scenarios (e.g., shopping malls, metro stations, stadiums etc.). Thus, developing a single best model with top accuracy in all applications, using any surveillance method, and in scenarios is not an effective solution. In fact, considering the existing trends in crowd counting model design, such a model will have large size, require huge compute resource for fine tuning, and incur longer delays during inference. Such a model will not be suitable for applications with limited on-chip memory (in edge devices), battery-powered devices, and real-time inference requirement. Thus we envision and also witnessed some recent efforts to have application specific model designing i.e., lightweight models for real-time applications on resource-constrained devices, whereas dense models for optimum accuracy over dense crowds in server-based platforms.

\item \textit{\textbf{CNNs versus transformers?}}
Although transformers have shown breakthrough performance in applications like NLP and other sequence-to-sequence modelling, their performance in computer vision is not comparably the same as in the other domains. Specifically, in crowd counting tasks, transformers so far have limited accuracy improvements over large benchmark datasets. Moreover, transformer models for crowd counting available to date are relatively deep models. This is because of the use of pretrained ViT which has 86M parameters (base model with 12 layers) as compared to CNN-based models (e.g., CSRNet has 16M parameters). We believe transformer models yet have to bring substantial accuracy gains and efficiency improvement to complement CNN models.

\end{itemize}
\section*{Acknowledgement}
This publication was made possible by the PDRA award PDRA7-0606-21012 from the Qatar National Research Fund (a member of The Qatar Foundation). The statements made herein are solely the responsibility of the authors.

\bibliographystyle{ieeetr} 
\bibliography{biblio}



\end{document}